\documentclass{article}

\usepackage{microtype}
\usepackage{graphicx}
\usepackage{booktabs} 
\usepackage{subcaption}
\usepackage{caption}
\usepackage{float}
\usepackage{colortbl}

\usepackage{hyperref}

\usepackage[accepted]{icml2025}

\usepackage{amsmath}
\usepackage{amssymb}
\usepackage{mathtools}
\usepackage{amsthm}
\usepackage{multirow}

\usepackage[capitalize,noabbrev]{cleveref}

\theoremstyle{plain}

\theoremstyle{definition}

\theoremstyle{remark}

\usepackage[textsize=tiny]{todonotes}

\icmltitlerunning{Explicit Temporal Modeling in Multimodal Large Language Models}

\begin{document}

\twocolumn[
\icmltitle{Exploring the Role of Explicit Temporal Modeling in Multimodal Large Language Models for Video Understanding}



\icmlsetsymbol{equal}{*} 

\begin{icmlauthorlist}
\icmlauthor{Yun Li}{equal,yyy} 
\icmlauthor{Zhe Liu}{equal,comp} 
\icmlauthor{Yajing Kong}{comp}
\icmlauthor{Guangrui Li}{comp}
\icmlauthor{Jiyuan Zhang}{comp}
\icmlauthor{Chao Bian}{comp}
\icmlauthor{Feng Liu}{sch}
\icmlauthor{Lina Yao}{yyy}
\icmlauthor{Zhenbang Sun}{comp}
\end{icmlauthorlist}

\icmlaffiliation{yyy}{CSIRO Data61, Australia}
\icmlaffiliation{comp}{TikTok, Australia}
\icmlaffiliation{sch}{The University of Melbourne, Australia}

\icmlcorrespondingauthor{Yun Li}{y.li@csiro.au}
\icmlkeywords{Multimodal Large Language Model, Temporal Learning}

\vskip 0.3in
]



\printAffiliationsAndNotice{\icmlEqualContribution} 

\begin{abstract}
Applying Multimodal Large Language Models (MLLMs) to video understanding presents significant challenges due to the need to model temporal relations across frames. Existing approaches adopt either \textit{implicit temporal modeling}, relying solely on the LLM decoder, or \textit{explicit temporal modeling}, employing auxiliary temporal encoders. To investigate this debate between the two paradigms, we propose the Stackable Temporal Encoder (STE). STE enables flexible explicit temporal modeling with adjustable temporal receptive fields and token compression ratios. Using STE, we systematically compare implicit and explicit temporal modeling across dimensions such as overall performance, token compression effectiveness, and temporal-specific understanding. We also explore STE's design considerations and broader impacts as a plug-in module and in image modalities. Our findings emphasize the critical role of explicit temporal modeling, providing actionable insights to advance video MLLMs.
\end{abstract}

\section{Introduction }

\begin{figure}[ht!]
    \centering
    \includegraphics[width=\linewidth]{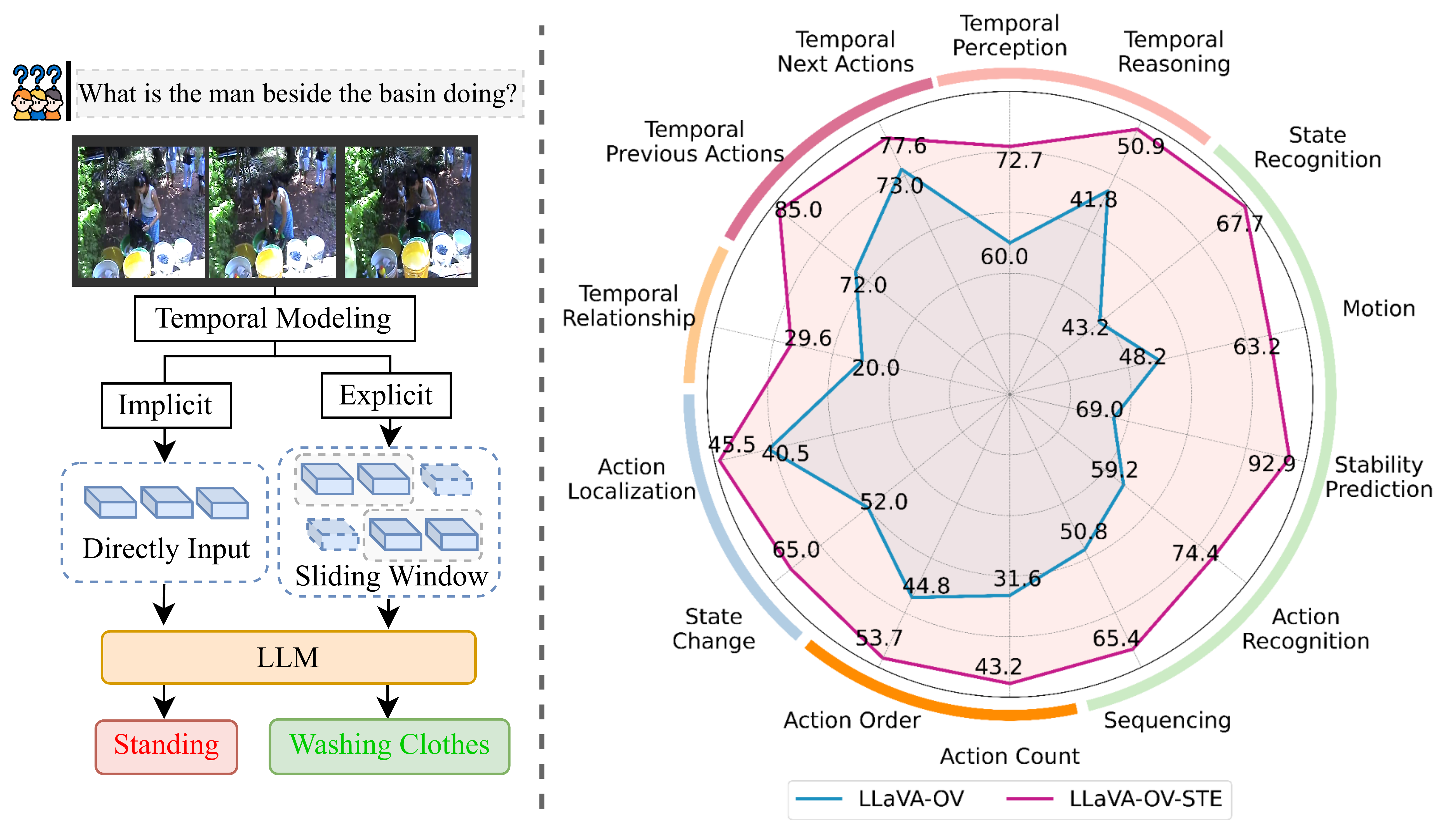}
    \caption{(Left): Explicit temporal modeling may enhance temporal understanding compared to implicit temporal modeling. (Right): Performance on temporal-related tasks of LLaVA-OV with (labeled as STE) or without explicit temporal modeling across six benchmarks (arc colors indicate different benchmarks).}
    \label{fig:intro}
\end{figure}

Large Language Models (LLMs)~\cite{openai2022chatgpt, ouyang2022training, touvron2023llama,jiang2023mistral} have revolutionized AI with their remarkable ability to understand and follow human instructions, excelling in various tasks across domains. Building on these capabilities, MLLMs extend LLMs to visual modalities, enabling them to process and interpret images effectively~\cite{achiam2023gpt,bai2023qwen,liu2024visual,dong2024internlm}.
However, extending MLLMs from static images to videos poses a significant challenge due to the need for temporal understanding. Videos innately require understanding dynamic relationships across frames to model time dependencies, which is essential for tasks such as temporal reasoning.

Recently, increasing attention has been paid to improving video understanding in MLLMs~\cite{zhang2023video,li2023videochat,team2024gemini,ren2024timechat,shen2024longvu}.
These methods vary in how they model temporal information, leading to two contrasting design philosophies. 
\textbf{Relying only on the LLM Decoder:} 
Some MLLMs leave temporal understanding entirely to the LLM decoder~\cite{xu2024pllava,liu2024llavanext,ataallah2024minigpt4,li2024llava,zhang2024video,liu2025st}. They depend on the LLM to implicitly infer temporal relationships from sequential visual features. 
\textbf{Incorporating Explicit Temporal Modeling:} 
The other stream, in contrast, considers them as insufficient and 
introduces temporal encoders to explicitly model temporal dependencies before passing extracted features to the LLM. These encoders~\cite{liu2024oryx,liu2024kangaroo,li2025llama,cheng2024videollama} can employ temporal aggregation or compression techniques.

This distinction mirrors an early debate in LLMs regarding the decoder-only or encoder-decoder structures \cite{fu2023decoder,nielsen2024encoder}. In video temporal understanding, 
as shown in Fig.~\ref{fig:intro} (Left), relying solely on the LLM may process frames independently, failing to capture temporal dependencies. In contrast, explicit temporal modeling, like sliding windows, encodes these dependencies, improving temporal understanding but increasing structure complexity. Although both design choices are gaining traction, systematic comparisons to discuss the necessity of explicit temporal modeling for video MLLMs are still lacking.

Addressing this question is critical for guiding the design of future video MLLMs. To bridge the gap, we propose a novel Stackable Temporal Encoder (STE). STE learns visual temporal information and offers adjustable temporal receptive fields and flexible token compression. By integrating STE into two SOTA open-source models from LLaVA series, \textit{LLaVA-OV}~\cite{li2024llava} and \textit{LLaVA-Video}~\cite{zhang2024video} (both adopts implicit temporal modeling), we systematically explore the necessity, design and impact of explicit temporal modeling in video MLLMs as follows:

\textbf{Do video MLLMs need explicit temporal modeling?} \\
Across six video benchmarks, we find that incorporating the proposed STE into LLaVA-OV and LLaVA-Video improves their performance by 4.7\% and 1.5\%, respectively. 
Additionally, incorporating our STE for explicit temporal modeling enables significant frame compression. Compressing frames by 87.5\% with LLaVA-OV-STE leads to a slight accuracy improvement of 0.8\%, while a 75\% frame compression with LLaVA-Video-STE results in a modest accuracy drop of 0.5\%, compared to their original versions with 32 frames.
These results demonstrate the effectiveness of explicit temporal modeling compared to implicit temporal modeling.

\textbf{Does explicit temporal modeling truly improve temporal understanding?} \\
We show in Fig.~\ref{fig:intro} (Right) LLaVA-OV's performance on fourteen temporal-related tasks across six benchmarks. All tasks exhibit significant performance gains after incorporating the proposed STE, validating that explicit temporal modeling can improve temporal understanding.

\textbf{How should we design explicit temporal modeling?}  \\
We investigate two key aspects of explicit temporal modeling: temporal receptive fields and learning space. Our findings reveal that simply increasing the temporal receptive fields does not substantially improve model performance with the fixed frame number. However, it can help mitigate information loss when compressing frames. Additionally, we observe that using STE to model temporal information explicitly is more effective in the visual space than in the semantic space. Specifically, applying convolutions before the vision-language projector yields better performance while requiring significantly fewer parameters.

\textbf{What are the broader implications?} \\
Without Supervised Fine-Tuning (SFT), STE acts as a lightweight plug-in module pre-trained on video datasets, achieving drops of less than 1.9\% when reducing 75\% frames for both models. These findings highlight its adaptability and potential for fast deployment in practical scenarios. Additionally, we examine its effects beyond video modalities: while single-image performance experiences a slight decline, likely due to the absence of image datasets during fine-tuning, multi-image tasks benefit significantly. 

In summary, we introduce STE to explore the role of explicit temporal modeling in video MLLMs. Our findings confirm its effectiveness, identify key design factors, highlight its compression benefits, and demonstrate its broader implications across modalities and potentials as a plug-in module. These findings provide insights for improving temporal understanding in future video MLLMs. We will release the codes recently.

\begin{figure*}[ht!]
    \centering
    \includegraphics[width=0.8\textwidth,keepaspectratio]{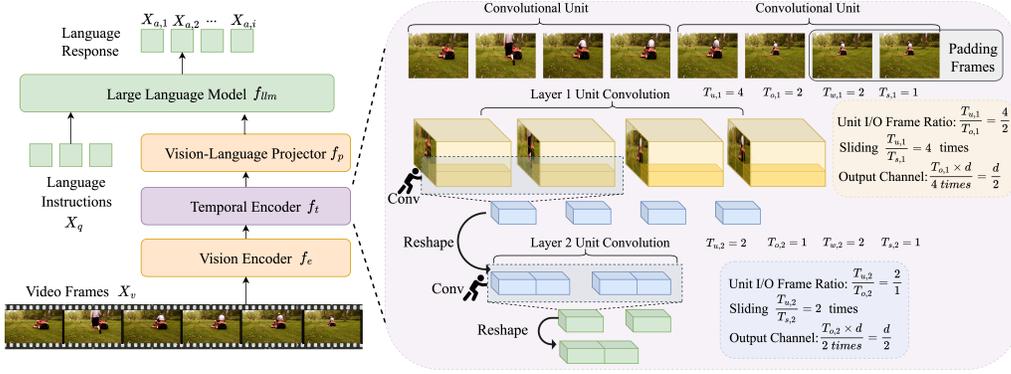}
    \caption{(Left) Overview of our model for processing video inputs. (Right) Schematic diagram of the temporal encoder, comprising 2-layer STE modules that encode every four frames into one abstract frame through stacking two layers of 50\% frame compression. The video, with dynamic length, is divided into convolutional units, and the STE is designed to handle diverse Input/Output (I/O) frame ratios based on these units. $T_{u,l}$, $T_{o,l}$, $T_{w,l}$, and $T_{s,l}$ denote the input frame count, the target output frame count, the convolutional window size, and the convolutional stride for a convolutional unit in the $l$-th layer, respectively.}
    \label{fig:model_overview}
\end{figure*}

\section{Related work}
LLMs~\cite{touvron2023llama,alpaca,chiang2023vicuna}, such as ChatGPT~\cite{openai2022chatgpt}, GPT-4~\cite{achiam2023gpt}, and Claude~\cite{anthropic2024claude}, have demonstrated remarkable performance across various language tasks.
Building on the success of LLMs, MLLMs have emerged, integrating visual and textual modalities to enable advanced image-text understanding~\cite{alayrac2022flamingo, driess2023palm,zhu2023minigpt}. MLLMs such as LLaVA~\cite{liu2023llava} leverage instruction tuning and visual-language alignment to achieve SOTA performance.

Video MLLMs extend image MLLMs to handle video inputs~\cite{li2023videochat,team2024gemini,ren2024timechat,shen2024longvu}, raising the challenge of modeling temporal dynamics. Earlier approaches such as VideoChat~\cite{li2023videochat} and Valley~\cite{luo2023valley} rely on static visual encoders such as CLIP~\cite{radford2021learning} and pooling strategies~\cite{xu2024pllava,liu2024llavanext} to compress frame features into compact representations, which are then passed to the LLM and rely on the LLM decoder to learn temporal information implicitly. Similarly, more recent models such as ST-LLM~\cite{liu2025st} and our baselines LLaVA-OV~\cite{li2024llava} and LLaVA-Video~\cite{zhang2024video} also belong to this implicit temporal modeling without delicately designed temporal modeling. 

However, other approaches~\cite{liu2024oryx,liu2024kangaroo,li2025llama,cheng2024videollama} highlight that implicit temporal modeling may limit temporal understanding and adopt explicit temporal learning modules. For example, Kangaroo~\cite{liu2024kangaroo}  and VideoLLaMA2~\cite{cheng2024videollama} introduce spatiotemporal convolutional connectors, and TimeChat~\cite{ren2024timechat} combines a timestamp-aware encoder with a sliding Q-Former. Temporal compression methods further improve efficiency; LLaMA-VID~\cite{li2025llama} reduces each frame to two tokens, while Oryx employs a dynamic temporal compressor~\cite{liu2024oryx}.

\textbf{Research Gap and Our Contributions}. While implicit and explicit temporal modeling strategies are widely used, there have been no systematic comparisons between them, and research on designing explicit temporal modeling remains limited. Existing explicit temporal modeling methods improve temporal understanding but often rely on fixed temporal learning scales~\cite{ren2024timechat} or compression ratios~\cite{li2025llama,liu2024oryx}, limiting their flexibility to explore how these factors affect model performance.

To address these gaps, we propose a novel explicit temporal learning module, the Stackable Temporal Encoder (STE), to facilitate comparisons between the two temporal modeling strategies. STE preserves temporal order through local convolution processing while allowing adjustable receptive fields to explore various temporal learning scales. Moreover, its flexible frame Input/Output (I/O) ratio accommodates both variable frame compression ratios and uncompressed outputs. This adaptability enables systematic studies of explicit temporal modeling and its broader impact.

\section{Methodology}

To explore the role of explicit temporal encoding in video MLLMs, we build upon the established open-source LLaVA series~\cite{liu2023llava}, preserving its core paradigm while integrating a novel Stackable Temporal Encoder (STE). This encoder is specifically designed to capture visual temporal information, facilitating a systematic investigation into what information temporal encoding learns and how the learned information influences the overall model. 

\subsection{Overall Architecture}

The overall architecture is shown in Fig.~\ref{fig:model_overview} (Left), illustrating how we integrate our STE module with the LLaVA architecture. Given visual inputs \( X_{v} \) and an instruction \( X_{q} \), we employ a vision encoder to extract visual embeddings \( Z_{v}=f_{e}(X_{v}) \). Our STE is then applied to capture temporal information within these embeddings, resulting in \( Z^{'}_{v}=f_{t}(Z_{v}) \). The vision-language projector will further project visual embedding \( Z^{'}_{v} \) into the semantic space, generating semantic tokens \( H_{v}=f_{p}(Z^{'}_{v}) \). These semantic tokens are combined with the instruction tokens and any previously generated tokens, and are fed into an LLM decoder to predict the next output token \( x_{i} = f_{llm}(H_{v}, X_{q}, X_{a,<i}) \), where \( X_{a,<i} \) denotes the answer tokens responding to the instruction generated before \( x_{i} \).
Then, we can compute the probability of the target answers \( X_a \) in a causal manner as:
\begin{equation}
    p(X_a | X_v, X_q) = \prod_{i=1}^L p(x_i | X_v, X_q, X_{a,<i})
\end{equation}

\subsection{Stackable Temporal Encoder (STE)}\label{sec:ste}

In this section, we introduce our novel temporal encoder, STE, which encodes temporal information through convolutions across frames and enables flexible compression ratios by adjusting the number of output convolution channels. Each output channel of the convolutional layers represents a distinct view of the input frames within the sliding windows. By varying the output channel count, the convolutional layer learns multiple views of temporally continuous information, encoding temporal dependencies and generating embeddings with the desired output shape. This design supports an arbitrary Input/Output (I/O) frame ratio of (\(\mathbb{N}_{+}\):\(\mathbb{N}_{+}\)), enabling STE to transform any number of input frames into any number of output frames, thereby facilitating flexible encoding and compression explorations.

As shown in Fig.~\ref{fig:model_overview} (Right), given a sequence of visual embeddings \(Z_{t}\) for \(t \in [1, T]\), where \(Z_{t} \in \mathbb{R}^{1 \times p \times d}\) represents the embedding of a frame, with \(T\) as the number of frames and \(p\) as the number of patches per frame, STE first concatenates these embeddings along the temporal dimension to enable temporal convolution:
\begin{equation}
    Z_{v} = [Z_{1}, Z_{2}, \dots, Z_{T}], \quad Z_{v} \in \mathbb{R}^{1 \times p \times (dT)}.
\end{equation}

Since the input frame length is dynamic, it is intractable to directly calculate the output channel count to implement the arbitrary I/O frame ratio of (\(N_{+}\):\(N_{+}\)). Therefore, we split the dynamic video into multiple convolutional units. In each unit, we perform the same operations to ensure the same I/O ratio within the unit and, thus, for the whole dynamic video.

We define (\(T_{u,l}:T_{o,l}\)) as the frame I/O ratio for the convolutional unit in the \(l\)-th layer of STE, where \(T_{u,l} \in \mathbb{N}_{+}\) denotes the input frame count for each convolutional unit, and \(T_{o,l} \in \mathbb{N}_{+}\) represents the target output frame count after compression. We then pad the dynamic-length video as:
\begin{equation}
Z_{v} = [Z_{1}, Z_{2}, \dots, Z_{T}, \underbrace{Z_{T}, Z_{T}, \dots, Z_{T}}_{k \text{ times}}],
\end{equation}
where \(k = T_{u,l} - (T \mod T_{u,l})\). We pad the video for two reasons: enabling the video length to be split into multiple convolutional units, and ensuring that the video can be compressed at a ratio of (\(T_{u,l}:T_{o,l}\)). For example, a video with 31 frames cannot be directly processed with a ratio of (2:1) along the temporal dimension unless 1 frame is padded.

We further annotate the frames included in a convolution window and a stride as \( T_{w,l} \) and \( T_{s,l} \). To run our sliding mechanism successfully, we need to ensure: \( T_{w,l} \leq T_{u,l} \), and \( T_{u,l} \mod T_{s,l} = 0 \). 
These conditions guarantee consistent operations for each convolutional unit across the entire video. Then, the output channel count \( C_{l} \) can be calculated as \(C_{l} = \frac{T_{o,l} \times d}{N}\),
where \( N = \frac{T_{u,l}}{T_{s,l}} \) denotes the number of sliding operations in each convolutional unit. Circular padding is applied to ensure that exactly \( N \) sliding operations can be performed for any \( T_{w,l} \).

Using the designed output channel count \( C_{l} \), the desired frame I/O  ratio \((T_{u,l}, T_{o,l})\) for the \( l \)-th convolutional layer can be achieved by concatenating the outputs of different channels along the temporal dimension, i.e., in the sliding order. The total output shape of the entire video is:
\[
Z_{v,l}' \in \mathbb{R}^{1 \times p \times (\frac{T + k}{T_{u,l}} \times N \times C_{l})} = \mathbb{R}^{1 \times p \times \frac{(T + k)T_{o,l}d}{T_{u,l}}},
\]
where \( Z_{v,l}' \) denotes the output of the \( l \)-th STE layer. 

The visual embedding for each patch can then be split into \( \frac{(T + k)T_{o,l}}{T_{u,l}} \) pieces of \( d \)-dimensional embeddings, each representing an abstract encoded frame. This ensures size compatibility when stacking STE layers. Thus, the final output \( Z_{v}' \) for the vision-language projector is expressed as:
\begin{equation}
    Z_{v}' = \underbrace{f_{t,L}(f_{t,L-1}(\dots(f_{t,1}}_{L \text{ layers}}(Z_{v})))),
\end{equation}
where \( L \) is the number of stacked layers in the STE \( f_{t} \).

\subsection{Training}
Our training process consists of two stages: Pretraining and Supervised Fine-Tuning (SFT). The training paradigm in both stages follows the conventional LLaVA framework.

\textbf{Pretraining Stage}: This stage utilizes the backbone's video understanding capabilities to initialize the STE, enabling it to effectively capture temporal information in videos. During pretraining, we only train the STE \( f_{t}(\cdot) \) while keeping the rest of the model (i.e., \( f_{e}(\cdot) \), \( f_{p}(\cdot) \), and \( f_{\text{llm}}(\cdot) \)) frozen. In this stage, we adopt a relatively high learning rate of \( 1 \times 10^{-3} \) to learn temporal information.

\textbf{SFT Stage}: In this stage, the entire model is fine-tuned to enable a comprehensive understanding of visual inputs with encoded temporal information. Smaller learning rates are applied: \( 2 \times 10^{-6} \) for the vision encoder and \( 1 \times 10^{-5} \) for the remaining components of the model.

\section{Experiment}

\subsection{Experiment Setting}

We follow the training and evaluation paradigms established by the LLaVA series~\cite{zhang2024video}. Specifically, we adopt Siglip-so400m~\cite{zhai2023sigmoid} as the vision encoder, Qwen2~\cite{qwen2} as the LLM decoder, and a two-layer fully connected network (FCN) as the vision-language projector. We load pre-trained parameters from LLaVA-OV-7B~\cite{li2024llava} and LLaVA-Video-7B~\cite{zhang2024video} as backbones to investigate the temporal modeling capabilities of STE. We use the LMMs-Eval framework the backbones adopt to evaluate our model and reproduce the backbone results for fair comparisons.
We use GPT-4o for GPT-based evals when the default GPT version in LMMs-Eval is unavailable. 

\begin{table*}[ht]
\caption{
Model performance comparisons. The performance of backbones integrated with our STE is shaded in grey.
The setting (2:2) represents (\(T_u\):\(T_o\)) for a specific STE layer. 
The best and second-best scores of open-source models are in \textbf{bold} and \underline{underlined}. }
\label{main_exp}
\centering
\resizebox{\textwidth}{!}{%
\begin{tabular}{c|l|cc|ccccccc}
\toprule
 & \multirow{2}{*}{Model} & \multirow{2}{*}{Size} & \multirow{2}{*}{\# Frames} & \multirow{2}{*}{PerceptionTest} & \multirow{2}{*}{ActNet-QA} & \multirow{2}{*}{NExT-QA} & \multirow{2}{*}{MLVU} & \multirow{2}{*}{MVBench} & VideoMME & VideoMME \\ 
 & & &  & & &  & & & (w/o sub) & (w/sub) \\ 
 \midrule
 & GPT-4V~\cite{openai2023gpt4v} & - & - & - & 57.0 & - & 49.2 & 43.5 & 59.9 & 63.3 \\ 
 & GPT-4o~\cite{openai2024hellogpt4o} & - & - & - & - & - & 64.6 & - & 71.9 & 77.2 \\ 
 & Gemini-1.5-Flash~\cite{team2024gemini} & - & - & - & 55.3 & - & - & - & 70.3 & 75.0 \\ 
\multirow{-4}{*}{Proprietary Models} & Gemini-1.5-Pro~\cite{team2024gemini} & - & - & - & 57.5 & - & - & - & 75.0 & 81.3 \\ 
\midrule
& \textbf{Implicit temporal encoding} &&&&&&&&&\\
& Video-LLaVA~\cite{lin2023video} & 7B & 8 & 44.3 & 45.3 & - & - & 41.0 & 39.9 & 42.6 \\ 
& LLaVA-N-Video~\cite{liu2024llavanext} & 7B & 32 & 48.8 & 53.5 & - & - & 46.5 & - & - \\ 
& LLaVA-N-Video~\cite{liu2024llavanext} & 32B & 32 & 59.4 & 54.3 & 77.3 & 65.5 & - & 60.2 & 63.0 \\ 
& PLLaVA~\cite{xu2024pllava} & 34B & 16 & - & 60.9 & - & - & 58.1 & - & - \\ 
  & LLaVA-OV~\cite{li2024llava} & 7B & 32 & 57.1 & 58.1 & 79.4 & 65.2& 56.7 & 58.5 & 61.1 \\ 
\multirow{0}{*}{Open-source Models} & LLaVA-Video~\cite{zhang2024video} & 7B & 32 & 66.6 & 64.5 & 79.8 & 66.9& 57.7 & \underline{62.0} & \underline{64.4} \\ 
& \textbf{Explicit temporal encoding} &&&&&&&&&\\
& LLaMA-VID~\cite{li2025llama} & 7B & 1 fps & 44.6 & 47.4 & - & - & 41.9 & 25.9 & - \\

 & VideoLLaMA2~\cite{cheng2024videollama} & 7B & 16 & 51.4 & 50.2 & - & - & 54.6 & 47.9 & 50.3 \\ 
 & VideoLLaMA2~\cite{cheng2024videollama} & 72B & 16 & 57.5 & 55.2 & - & 61.2 & 62.0 & 61.4 & 63.1 \\ 
  & Kangaroo~\cite{liu2024kangaroo} & 8B & 64 & - & - & - & 61.0 & 61.1 & 56.0 & 57.6 \\
 & Oryx~\cite{liu2024oryx} & 7B & 128 & 68.6 & - & 81.9 & 67.5 & \underline{63.9} & 58.3 & 62.6 \\
& Oryx-1.5~\cite{liu2024oryx} & 7B & 128 & 70.0 & - & 81.8 & 67.5 & \textbf{67.6} & 58.8 & 64.2 \\

 \midrule
 & LLaVA-OV-STE &&&&&&&&&\\
 & \cellcolor{gray!30}  (2:2) & \cellcolor{gray!30}7B & \cellcolor{gray!30}32 & \cellcolor{gray!30}70.1 & \cellcolor{gray!30}\underline{65.7} & \cellcolor{gray!30}82.4 & \cellcolor{gray!30}66.9& \cellcolor{gray!30}57.8 & \cellcolor{gray!30}60.0 & \cellcolor{gray!30}63.1 \\ 
 & \cellcolor{gray!30}  (2:2)-(2:2) & \cellcolor{gray!30}7B & \cellcolor{gray!30}32 & \cellcolor{gray!30}70.5 & \cellcolor{gray!30}65.3 & \cellcolor{gray!30}\textbf{83.0} & \cellcolor{gray!30}67.2& \cellcolor{gray!30}57.9 & \cellcolor{gray!30}61.6 & \cellcolor{gray!30}63.7 \\ 
 & \cellcolor{gray!30}  (2:2)-(2:2)-(2:2) & \cellcolor{gray!30}7B & \cellcolor{gray!30}32 & \cellcolor{gray!30}70.6 & \cellcolor{gray!30}\textbf{65.8} & \cellcolor{gray!30}82.5 & \cellcolor{gray!30}66.8& \cellcolor{gray!30}57.7 & \cellcolor{gray!30}60.9 & \cellcolor{gray!30}63.8 \\ 
  \cline{2-11}
  & LLaVA-Video-STE &&&&&&&&&\\
 & \cellcolor{gray!30}(2:2) & \cellcolor{gray!30}7B & \cellcolor{gray!30}32 & \cellcolor{gray!30}\underline{72.1} & \cellcolor{gray!30}65.1 & \cellcolor{gray!30}\underline{82.8} & \cellcolor{gray!30}\textbf{68.9}& \cellcolor{gray!30}57.9 & \cellcolor{gray!30}\underline{62.0} & \cellcolor{gray!30}63.7 \\ 
 & \cellcolor{gray!30}(2:2)-(2:2) & \cellcolor{gray!30}7B & \cellcolor{gray!30}32 & \cellcolor{gray!30}\textbf{72.3} & \cellcolor{gray!30}65.6 & \cellcolor{gray!30}82.4 & \cellcolor{gray!30}\underline{68.1}& \cellcolor{gray!30}57.8 & \cellcolor{gray!30}61.1 & \cellcolor{gray!30}62.9 \\ 
\multirow{-8}{*}{STE} & \cellcolor{gray!30}(2:2)-(2:2)-(2:2) & \cellcolor{gray!30}7B & \cellcolor{gray!30}32 & \cellcolor{gray!30}71.9 & \cellcolor{gray!30}65.4 & \cellcolor{gray!30}82.1 & \cellcolor{gray!30}67.9& \cellcolor{gray!30}57.9 & \cellcolor{gray!30}\textbf{62.1} & \cellcolor{gray!30}\textbf{64.9} \\ 
\bottomrule
\end{tabular}%
}
\end{table*}

\begin{table*}[ht]
\caption{
Model performance with (shaded in grey) and without STE when varying frame compression ratios. AVG is the average accuracy.
}
\label{reduce_frame_exp}
\centering
\resizebox{0.88\textwidth}{!}{%
\begin{tabular}{c|l|ccccccc|c}
\toprule
 & \multirow{2}{*}{Model \(+\) STE} & \multirow{2}{*}{PerceptionTest} & ActNet-QA & \multirow{2}{*}{NExT-QA} & \multirow{2}{*}{MLVU} & \multirow{2}{*}{MVBench} & VideoMME & VideoMME & \multirow{2}{*}{AVG} \\
 & &  & (Acc/Score) & & & & (w/o sub) & (w/sub) & \\
\midrule
 & LLaVA-OV & 57.1 & 57.7/2.92 & 78.6 & 63.8& 54.4 & 57.2 & 60.8 & 61.4\\
 & \cellcolor{gray!30}\(+\) (4:3) & \cellcolor{gray!30}69.7 & \cellcolor{gray!30}60.3/3.03 & \cellcolor{gray!30}79.0 & \cellcolor{gray!30}67.0& \cellcolor{gray!30}55.8 & \cellcolor{gray!30}59.7 & \cellcolor{gray!30}62.4 & \cellcolor{gray!30}64.8\\
  \cline{2-10}
 & LLaVA-Video & 66.4 & 64.4/3.22 & 80.1 & 65.7& 57.7 & 61.9 & 64.4 & 65.8\\
\multirow{-4}{*}{-25\% Frames} & \cellcolor{gray!30}\(+\) (4:3) & \cellcolor{gray!30}69.4 & \cellcolor{gray!30}62.8/3.15 & \cellcolor{gray!30}80.7 & \cellcolor{gray!30}66.8& \cellcolor{gray!30}56.4 & \cellcolor{gray!30}61.0 & \cellcolor{gray!30}62.6 & \cellcolor{gray!30}65.7\\
\midrule
 & LLaVA-OV & 57.4 & 57.3/2.90 & 78.3 & 62.2& 54.4 & 55.8 & 59.2 & 60.7\\
 & \cellcolor{gray!30}\(+\) (4:3)-(4:3) & \cellcolor{gray!30}67.3 & \cellcolor{gray!30}59.2/2.97 & \cellcolor{gray!30}77.6 & \cellcolor{gray!30}66.4& \cellcolor{gray!30}54.1 & \cellcolor{gray!30}59.2 & \cellcolor{gray!30}62.2 & \cellcolor{gray!30}63.7\\
  \cline{2-10}
 & LLaVA-Video & 66.1 & 63.9/3.19 & 79.9 & 63.7& 57.4 & 59.2 & 62.0 & 64.6\\
\multirow{-4}{*}{-43.75\% Frames} & \cellcolor{gray!30}\(+\) (4:3)-(4:3) & \cellcolor{gray!30}68.8 & \cellcolor{gray!30}61.0/3.05 & \cellcolor{gray!30}79.9 & \cellcolor{gray!30}64.7& \cellcolor{gray!30}55.5 & \cellcolor{gray!30}59.1 & \cellcolor{gray!30}62.3 & \cellcolor{gray!30}64.5\\
\midrule
 & LLaVA-OV & 57.3 & 57.9/2.92 & 78.2 & 62.5& 53.4 & 56.2 & 59.4 & 60.7\\
 & \cellcolor{gray!30}\(+\) (2:1) & \cellcolor{gray!30}69.4 & \cellcolor{gray!30}60.1/3.01 & \cellcolor{gray!30}79.1 & \cellcolor{gray!30}66.9& \cellcolor{gray!30}55.3 & \cellcolor{gray!30}60.4 & \cellcolor{gray!30}63.2 & \cellcolor{gray!30}64.9\\
  \cline{2-10}
 & LLaVA-Video & 65.9 & 63.4/3.17 & 79.5 & 63.9& 57.4 & 59.7 & 62.0 & 64.5\\
\multirow{-4}{*}{-50\% Frames} & \cellcolor{gray!30}\(+\) (2:1) & \cellcolor{gray!30}70.9 & \cellcolor{gray!30}62.5/3.13 & \cellcolor{gray!30}80.7 & \cellcolor{gray!30}67.8& \cellcolor{gray!30}57.6 & \cellcolor{gray!30}61.0 & \cellcolor{gray!30}63.9 & \cellcolor{gray!30}66.3\\
\midrule
 & LLaVA-OV & 56.5 & 56.4/2.84 & 77.4 & 59.3& 52.8 & 53.6 & 57.3 & 59.0\\
 & \cellcolor{gray!30}\(+\) (2:1)-(2:1) & \cellcolor{gray!30}67.9 & \cellcolor{gray!30}59.7/3.00 & \cellcolor{gray!30}79.0 & \cellcolor{gray!30}66.2& \cellcolor{gray!30}54.8 & \cellcolor{gray!30}59.4 & \cellcolor{gray!30}62.4 & \cellcolor{gray!30}64.2\\
  \cline{2-10}
 & LLaVA-Video & 64.4 & 62.1/3.01 & 76.1 & 58.9& 56.0 & 56.0 & 60.4 & 62.0\\
\multirow{-4}{*}{-75\% Frames} & \cellcolor{gray!30}\(+\) (2:1)-(2:1) & \cellcolor{gray!30}69.8 & \cellcolor{gray!30}61.6/3.09 & \cellcolor{gray!30}80.0 & \cellcolor{gray!30}66.6& \cellcolor{gray!30}57.9 & \cellcolor{gray!30}59.7 & \cellcolor{gray!30}62.8 & \cellcolor{gray!30}65.5\\
\midrule
 & LLaVA-OV & 55.6 & 53.6/2.71 & 76.1 & 56.1& 51.4 & 49.9 & 55.5 & 56.9\\
 & \cellcolor{gray!30}\(+\) (2:1)-(2:1)-(2:1) & \cellcolor{gray!30}67.1 & \cellcolor{gray!30}58.8/2.96 & \cellcolor{gray!30}77.7 & \cellcolor{gray!30}63.6& \cellcolor{gray!30}53.6 & \cellcolor{gray!30}58.7 & \cellcolor{gray!30}62.2 & \cellcolor{gray!30}63.1\\
  \cline{2-10}
 & LLaVA-Video & 62.6 & 58.5/2.93 & 74.3 & 56.2& 53.8 & 52.2 & 57.6 & 59.3\\
\multirow{-4}{*}{-87.5\% Frames} & \cellcolor{gray!30}\(+\) (2:1)-(2:1)-(2:1) & \cellcolor{gray!30}68.6 & \cellcolor{gray!30}60.7/3.04 & \cellcolor{gray!30}79.9 & \cellcolor{gray!30}64.3& \cellcolor{gray!30}55.4 & \cellcolor{gray!30}59.4 & \cellcolor{gray!30}62.0 & \cellcolor{gray!30}64.3\\
\midrule
 & LLaVA-OV & 54.6 & 44.0/2.22 & 73.6 & 51.9& 48.8 & 43.6 & 51.8 & 52.6\\
 & \cellcolor{gray!30}\(+\) (2:1)-(2:1)-(2:1)-(2:1) & \cellcolor{gray!30}61.6 & \cellcolor{gray!30}54.4/2.74 & \cellcolor{gray!30}74.5 & \cellcolor{gray!30}61.0& \cellcolor{gray!30}50.2 & \cellcolor{gray!30}53.1 & \cellcolor{gray!30}58.0 & \cellcolor{gray!30}59.0\\
  \cline{2-10}
 & LLaVA-Video & 60.8 & 48.0/2.42 & 71.0 & 51.7& 50.5 & 47.0 & 53.4 & 54.6\\
\multirow{-4}{*}{-93.75\% Frames} & \cellcolor{gray!30}\(+\) (2:1)-(2:1)-(2:1)-(2:1) & \cellcolor{gray!30}66.7 & \cellcolor{gray!30}58.2/2.92 & \cellcolor{gray!30}78.3 & \cellcolor{gray!30}61.3& \cellcolor{gray!30}54.4 & \cellcolor{gray!30}57.5 & \cellcolor{gray!30}60.8 & \cellcolor{gray!30}62.5\\
\bottomrule
\end{tabular}%
}
\end{table*}

To explore the role of explicit temporal modeling, we vary \((T_{u}\):\(T_{o})\) across different scenarios. When \((T_{u}\):\(T_{o})\) is set to (2:2), the frame count remains fixed, allowing us to study the influence of temporal encoding without frame compression. For \((T_{u}\):\(T_{o})\) configurations of (4:3) and (2:1), we examine the model's performance under layer-wise compression ratios of 25\% and 50\%, respectively. Stacking these layers multiple times can reduce from 25\% to 93.75\% frames. This enables us to investigate how explicit temporal modeling and compression ratios affect model performance. Other experiment details can be found in Appendix~\ref{exp_set_details}.

\subsection{Training and Evaluation Datasets}
We train the model using the video training data of LLaVA-Video, excluding the 1.1 million image-language pairs. This allows us to focus on the video datasets. Details of the training datasets, i.e., ActNet-QA~\cite{yu2019activitynet}, {NExT-QA}~\cite{xiao2021next}, {PerceptionTest}~\cite{patraucean2023perception}, {LLaVA-Hound}~\cite{zhang2024direct}, and {LLaVA-Video-178K}~\cite{zhang2024video}, are in Appendix \ref{training_dataset_details}.

We evaluate STE on six open-ended and multiple-choice video benchmarks: ActNetQA~\cite{yu2019activitynet} and NExT-QA~\cite{xiao2021next}, focusing on spatio-temporal reasoning of activities and actions; PerceptionTest~\cite{patraucean2023perception}, testing perception ability; MLVU~\cite{zhou2024mlvu}, emphasizing long video understanding; and VideoMME~\cite{fu2024video} and MVBench~\cite{li2024mvbench}, offering comprehensive evaluations. 

Additionally, we assess STE on image benchmarks to analyze its impact on image understanding abilities after SFT. These include single-image dataset: AI2D~\cite{kembhavi2016diagram}, DocVQA~\cite{mathew2021docvqa}, InfoVQA~\cite{mathew2022infographicvqa}, MMMU~\cite{yue2024mmmu}, MME~\cite{yin2023survey}, MMBench~\cite{liu2025mmbench}, MMStar~\cite{chen2024we}, and RealworldQA~\cite{grok-1.5}; and multi-image dataset MuirBench~\cite{wang2024muirbench}.

\subsection{Effectiveness of Explicit Temporal Modeling}

We reveal \textbf{the necessity of explicit temporal modeling} by demonstrating its effectiveness in two scenarios: maintaining and compressing frame count. The results of varied STE structures after two-stage training (i.e., pretraining and SFT) are presented across six video benchmarks.

\textbf{Maintaining Frame Count}:  In this scenario (Tab.~\ref{main_exp}), we maintain the frame count and stack STE layers from 1 to 3 to explicitly model temporal information at different time scales. When comparing SOTA open-source video MLLMs, we observe that methods utilizing explicit temporal modeling, including ours, generally achieve better performance compared to implicit temporal modeling, particularly among models of similar sizes. This demonstrates the effectiveness of explicit temporal modeling in enhancing video understanding. More specifically, our method achieves an average performance improvement of up to 4.7\% and 1.5\% across six benchmarks compared to our backbones LLaVA-OV and LLaVA-Video, respectively. This demonstrates that explicitly temporal modeling significantly enhances the backbone's ability to understand video content.  A closer look at various benchmarks shows that our models consistently outperform the backbones on PerceptionTest, ActNet-QA, NExT-QA, MLVU, and MVBench, indicating that STE strengthens perception and reasoning capabilities. Moreover, our model outperforms VideoLLaMA2~\cite{cheng2024videollama} 72B, despite having only one-tenth the parameters, and Oryx, despite processing only one-fourth the input video frames. Notably, all three models—ours, VideoLLaMA2, and Oryx—employ explicit temporal modeling, highlighting that STE is more effective in enhancing temporal learning.

\begin{figure}[t]
    \centering
    \begin{subfigure}{0.495\linewidth}
        \centering
        \includegraphics[width=0.9\linewidth]{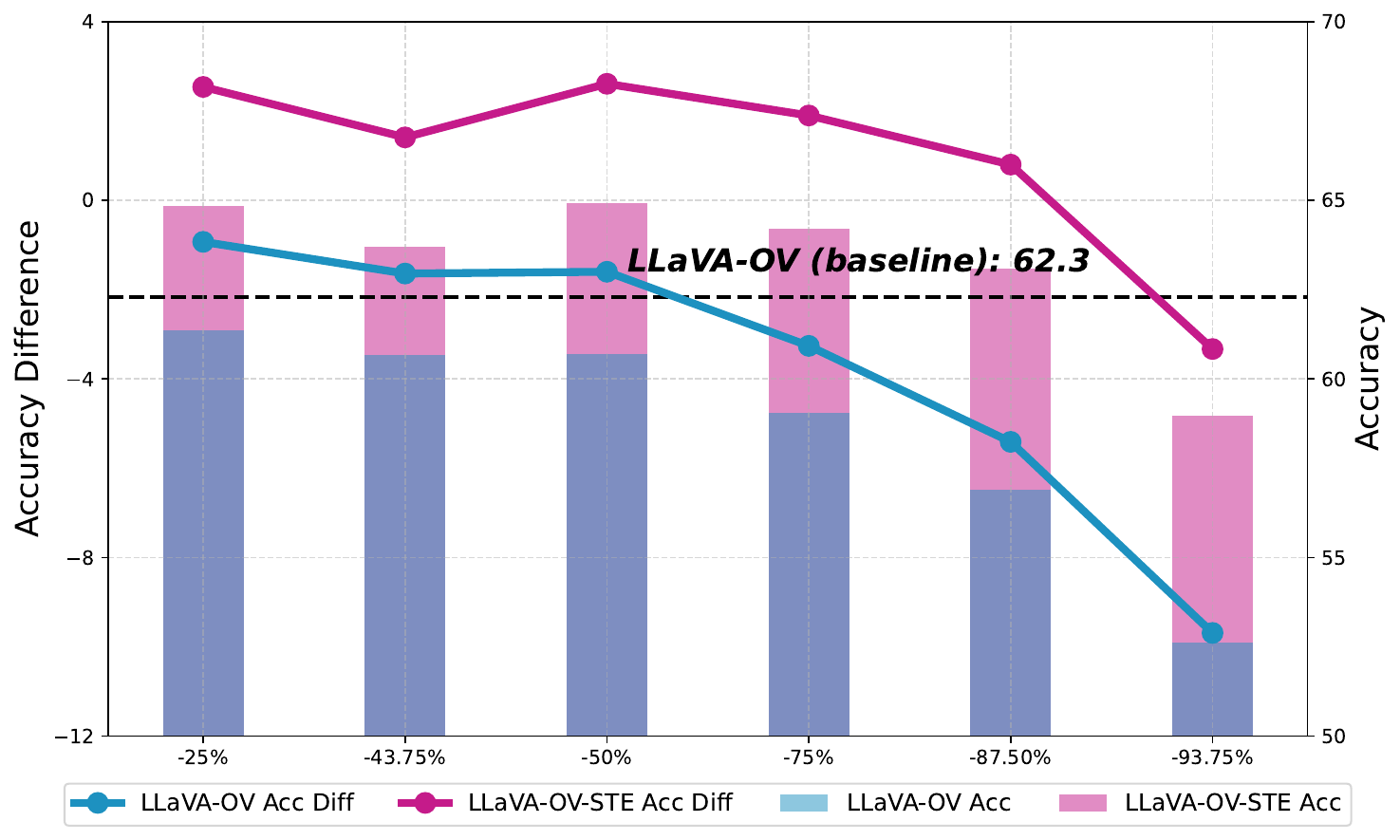}
        \caption{LLaVA-OV w or w/o STE.}
    \end{subfigure}%
    \hfill
    \begin{subfigure}{0.495\linewidth}
        \centering
        \includegraphics[width=0.9\linewidth]{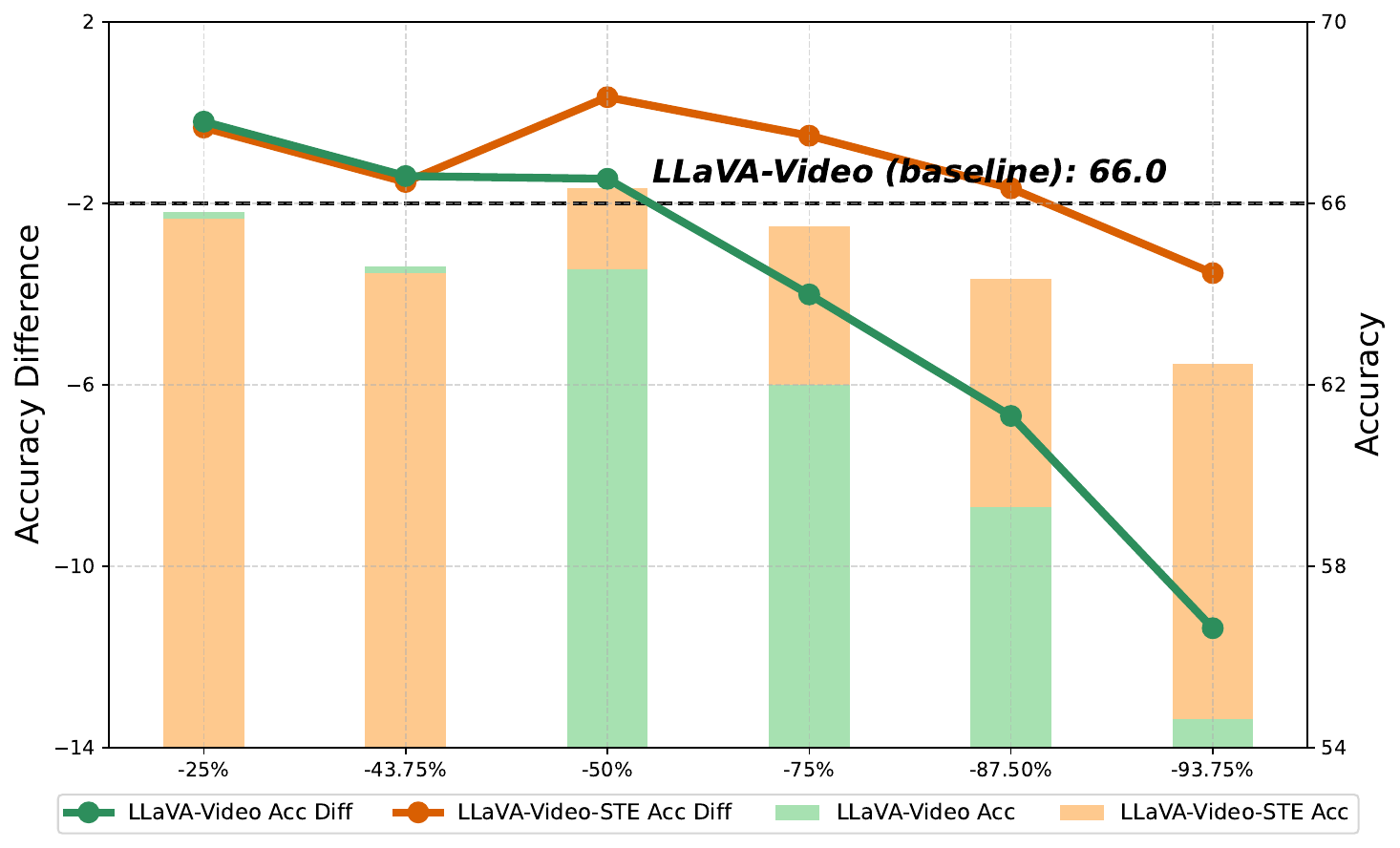}
        \caption{LLAVA-Video w or w/o STE.}
    \end{subfigure}
    \caption{Performance when varying frame compressions: sampling frequency reduction vs. frame compression (STE), showing accuracy differences relative to backbones with 32 input frames.}
    \label{fig:difference_curve}
\end{figure}

\begin{figure}[t]
\centering
    \begin{subfigure}{0.6\linewidth}
        \centering
        \includegraphics[width=\linewidth]{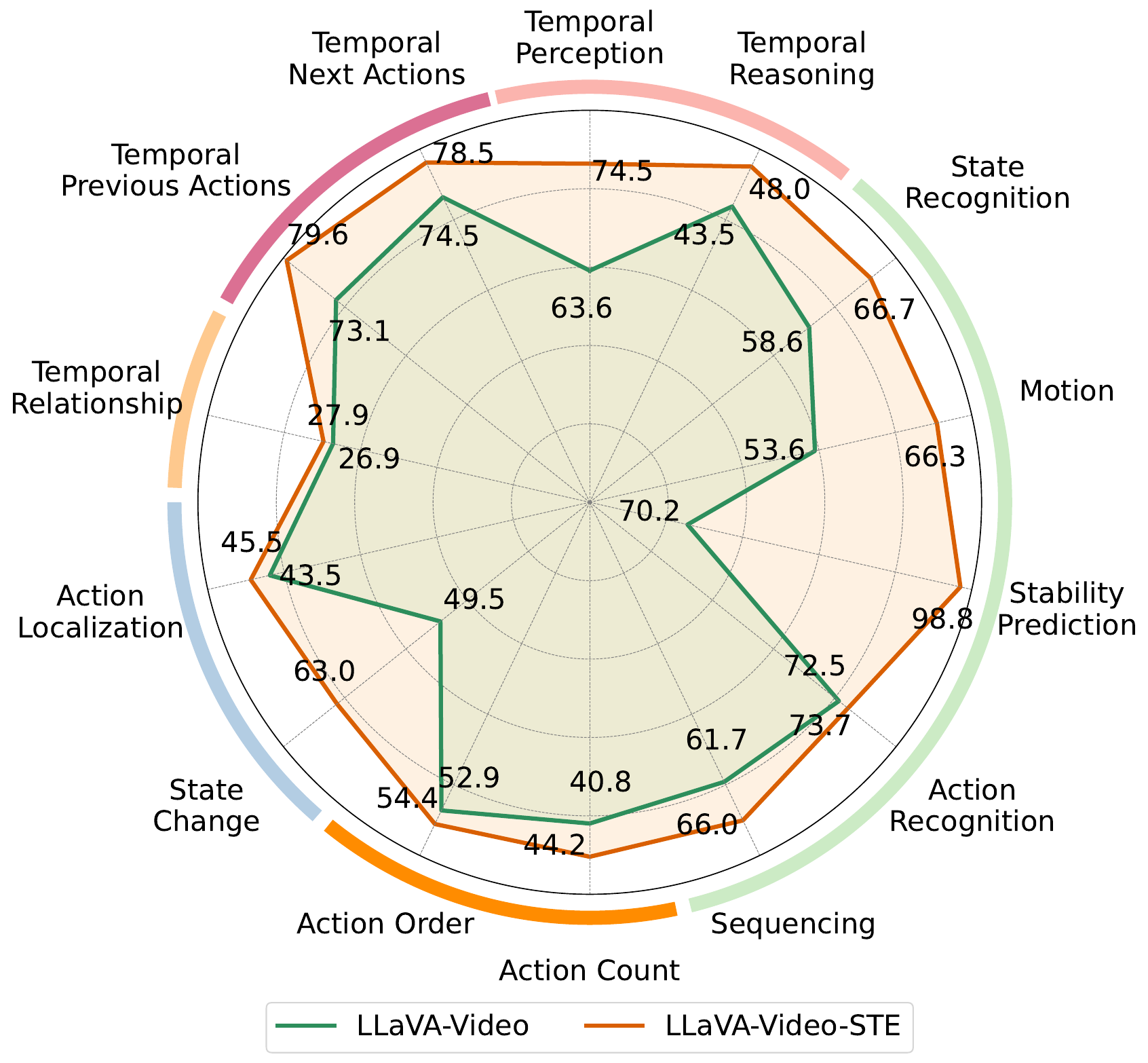}
    \end{subfigure}
    \caption{Performance of LLaVA-Video on temporal-related tasks equipped with (labeled as STE) or without explicit temporal modeling across benchmarks (arc colors indicate different benchmarks).}
    \label{fig:temporal_tasks_video}
\end{figure}

\textbf{Compressing Frame Count}: 
To compress frame counts, we stack STE layers with varied frame I/O ratios (\(T_u:T_o\)) for explicit temporal modeling or reduce the backbone's sampling frequency to lower input frame counts for implicit temporal modeling. Results for different frame compressions (e.g., -25\% frames = compression with (\(T_u:T_o\)) as (4:3), and -50\% frames = compression with (\(T_u:T_o\)) as (2:1)) are presented in Tab.~\ref{reduce_frame_exp} and visualized in Fig.~\ref{fig:difference_curve}, where bars represent accuracy and lines indicate accuracy differences relative to backbones with 32 input frames.

LLaVA-OV-STE consistently outperforms the baseline across frame compressions from 25\% to 87.5\%. LLaVA-Video-STE remains competitive, underperforming the baseline only at compressions beyond 75\%, with a small 0.5\% average decrease at 75\% compression. However, at 25\% and 43.75\% frame compressions, LLaVA-Video-STE performs worse than directly sampling fewer frames. These compressions correspond to 1 and 2 STE layers with frame I/O as (4:3), where merging the first frame and part of the second frame into a single abstract frame disrupts embedding completeness and information consistency. For reductions between 50\% and 93.75\%, STE-based frame compression demonstrates a slower performance decline compared to sampling fewer frames, highlighting its effectiveness in mitigating performance loss from reduced frame counts.

\begin{table*}[t]
\centering
\caption{Ablation study on temporal learning space: learning in visual or semantic space during pretraining and SFT stages.}
\label{tab:placement_ablation}
\resizebox{0.88\textwidth}{!}{%
\begin{tabular}{c|c|c|c|ccccccc|c}
\toprule
\multirow{2}{*}{Training Phase}       & \multirow{2}{*}{Backbone}       & \multirow{2}{*}{Before Projector}   & \multirow{2}{*}{After Projector}                 & \multirow{2}{*}{PerceptionTest} & \multirow{2}{*}{ActNet-QA} & \multirow{2}{*}{NExT-QA} & \multirow{2}{*}{MLVU} & \multirow{2}{*}{MVBench}& VideoMME & VideoMME & \multirow{2}{*}{AVG}   \\
                     &    & & & & & & & & (w/o sub) & (w/sub) &   \\
                     \midrule
 &  & (2:2) & $\times$ & 58.6 & 58.7 & 79.6 & 65.1 & 55.0 & 58.9 & 60.0 & 62.3 \\
 & \multirow{-2}{*}{LLaVA-OV} & $\times$ & (2:2) & 57.4 & 57.3 & 77.2 & 64.0 & 53.2 & 55.3 & 58.2 & 60.4 \\
 \cline{2-12}
 &  & (2:2) & $\times$ & 67.0 & 64.4 & 82.0 & 66.8 & 58.3 & 62.1 & 64.1 & 66.4 \\
\multirow{-4}{*}{Pretraining} & \multirow{-2}{*}{LLaVA-Video} & $\times$ & (2:2) & 65.4 & 63.2 & 81.0 & 65.7 & 55.5 & 60.7 & 63.7 & 65.0 \\
\midrule
 &  & (2:2) & $\times$ & 70.1 & 65.7 & 82.4 & 66.9 & 57.8 & 60.0 & 63.1 & 66.6 \\
 & \multirow{-2}{*}{LLaVA-OV} & $\times$ & (2:2) & 69.5 & 65.5 & 82.7 & 66.9 & 55.8 & 60.7 & 63.0 & 66.3 \\
  \cline{2-12}
 &  & (2:2) & $\times$ & 72.1 & 65.1 & 82.8 & 68.9 & 57.9 & 62.0 & 63.7 & 67.5 \\
\multirow{-4}{*}{SFT} & \multirow{-2}{*}{LLaVA-Video} & $\times$ & (2:2) & 71.4 & 65.2 & 82.2 & 67.2 & 56.5 & 61.7 & 63.1 & 66.8\\
\bottomrule

\end{tabular}%
}
\end{table*}

\begin{table*}[t]
\centering
\caption{Compression effectiveness when using STE as a plug-in module. }
\label{plug_in_exp}
\resizebox{0.88\textwidth}{!}{
\begin{tabular}{l|c|ccccccc|c}
\toprule
  & \multirow{2}{*}{Trainable Parameters} & \multirow{2}{*}{PerceptionTest} & \multirow{2}{*}{ActNet-QA} & \multirow{2}{*}{NExT-QA}  & \multirow{2}{*}{MLVU} & \multirow{2}{*}{MVBench} & VideoMME & VideoMME & \multirow{2}{*}{AVG} \\
  & & & & & & & (w/o sub) & (w/sub) &  \\
  \midrule
LLaVA-OV & N/A & 57.1 & 58.1 & 79.4 & 56.7 & 58.5 & 65.2 & 61.1 & 62.3\\
\midrule
-0\% frames & $\sim$2.65 M  & 58.6 & 58.7 & 79.6 & 65.1 & 55.0 & 58.9 & 60.0 & 62.3 \\
-50\% frames  & $\sim$1.33 M & 57.9 & 58.9 & 78.9  & 63.5 & 53.9 & 57.6 & 59.8 & 61.5\\
-75\% frames & $\sim$2.65 M & 56.9 & 57.2 & 78.1  & 63.0 & 53.6& 55.7 & 58.5 & 60.4\\
-87.5\% frames & $\sim$3.98 M & 55.7 & 56.2 & 76.8 & 60.5 & 52.1 & 53.9 & 57.4 & 58.9\\
-93.75\% frames  & $\sim$5.31 M & 46.4 & 45.2 & 67.2 & 55.9 & 40.1 & 45.6 & 51.6 & 50.3\\
\midrule
LLaVA-Video & N/A & 66.6 & 64.5 & 79.8  & 66.9 & 57.7 & 62.0 & 64.4 & 66.0\\
\midrule
-0\% frames & $\sim$2.65 M  & 67.0 & 64.4 & 82.0 & 66.8 & 58.3 & 62.1 & 64.1 & 66.4\\
-50\% frames & $\sim$1.33 M & 66.6 & 63.1 & 81.9  & 66.6 & 56.6 & 61.0 & 64.0 & 65.7\\
-75\% frames & $\sim$2.65 M & 65.1 & 62.4 & 80.3  & 64.4 & 55.7 & 59.5 & 62.1 & 64.2\\
-87.5\% frames & $\sim$3.98 M & 63.7 & 60.8 & 79.3 & 62.5 & 54.0 & 58.0 & 61.0 & 62.8\\
-93.75\% frames & $\sim$5.31 M & 61.3 & 58.3 & 77.3 & 60.2 & 52.9 & 55.2 & 59.3 & 60.6\\
\bottomrule
\end{tabular}
}
\end{table*}

\begin{table*}[]
\centering
\caption{Single-Image and Multi-image (MuirBench) Ability after SFT.}
\label{tab:image_ability}
\resizebox{0.88\textwidth}{!}{%
\begin{tabular}{l|ccccccccc|c|c}
\toprule
\multicolumn{1}{l|}{} & AI2D & DocVQA & InfoVQA & MMMU & MME-cog & MME-perp & RealWorldQA & MMBench & MMStar & MuirBench & Relative Score to OV(↑) \\
\midrule
LLaVA-OV & 81.4 & 87.5 & 68.8 & 48.8 & 418 & 1580 & 66.3 & 80.8 & 61.7 & 41.8 & 1.000 \\
\midrule
\(+\) (2:2) & 78.3 & 83.3 & 55.8 & 46.7 & 454 & 1435 & 56.9 & 80.6 & 58.9 & 45.2 & 0.957 \\
\(+\) (2:2)-(2:2) & 77.9 & 83.2 & 55.9 & 46.7 & 460 & 1432 & 55.6 & 81.4 & 57.8 & 45.3 & 0.955 \\
\(+\) (2:2)-(2:2)-(2:2) & 77.8 & 82.7 & 56.3 & 46.7 & 439 & 1441 & 57.7 & 81.5 & 57.9 & 43.2 & 0.949 \\
\(+\) (4:3) & 77.9 & 85.4 & 61.8 & 47.6 & 400 & 1416 & 61.7 & 80.3 & 58.3 & 45.9 & 0.963 \\
\(+\) (4:3)-(4:3) & 77.8 & 86.1 & 61.7 & 48.4 & 386 & 1475 & 63.1 & 80.1 & 58.3 & 45.3 & 0.966 \\
\(+\) (2:1) & 78.4 & 86.5 & 61.6 & 46.6 & 403 & 1413 & 59.4 & 81.2 & 59.3 & 46.5 & 0.963 \\
\(+\) (2:1)-(2:1) & 78.0 & 85.7 & 60.7 & 47.7 & 379 & 1370 & 62.6 & 81.1 & 58.3 & 46.4 & 0.957 \\
\(+\) (2:1)-(2:1)-(2:1) & 77.7 & 84.4 & 58.8 & 47.3 & 378 & 1425 & 61.6 & 80.4 & 57.6 & 45.9 & 0.950 \\
\(+\) (2:1)-(2:1)-(2:1)-(2:1) & 77.3 & 83.6 & 59.7 & 47.3 & 346 & 1337 & 60.7 & 78.8 & 55.7 & 43.9 & 0.926\\
\bottomrule
\end{tabular}
}
\end{table*}

\subsection{Evaluating Temporal Understanding Abilities}

We evaluate whether \textbf{explicit temporal modeling truly improves temporal understanding} by investigating task-level performance across six video benchmarks. We focus on tasks related to temporal understanding, comparing LLaVA-OV with and without a 3-layer (2:2) STE in Fig.\ref{fig:intro} and LLaVA-Video with and without a 3-layer (2:2) STE in Fig.\ref{fig:temporal_tasks_video}.
The results show that STE consistently improves temporal understanding across tasks, benchmarks, and backbones, particularly for stability prediction, motion recognition, temporal perception, and stage change. These findings demonstrate that explicitly modeling temporal information enables the model to better detect temporal changes and comprehend temporal dynamics, thereby enhancing its prediction, recognition, and perception capabilities over sequences of frames. Additional details on task-level performance are provided in Appendix \ref{task-level-ability}.

\textbf{Qualitative analysis.} We also present the qualitative results of LLaVA-Video with and without STE in Appendix~\ref{Qualitative} to explore how STE improves temporal understanding and compensates for information loss when compressing frames.

\subsection{STE Design Ablation}

\textbf{How should we design explicit temporal modeling?}  
To investigate this, we analyze two critical factors in the design of STE: the temporal receptive field and the temporal learning space. These analyses provide insights into how convolutional-based temporal encoders should be designed when explicit temporal modeling is needed. 

\textbf{Temporal receptive field.}  
We evaluate how expanding the temporal receptive field affects model performance by stacking 1, 2, and 3 STE layers of (2:2). This gradually increases the temporal receptive field covered by the sliding window in the final layer, while fixing all other parameters and retaining the full number of frames without compression.

As shown in Tab.~\ref{main_exp}, expanding the temporal receptive field does not lead to significant changes in overall performance, with average accuracy remaining relatively stable across configurations. This robustness can be attributed to the design of each STE layer: as we use a window size of 2 and a stride of 1, even a single layer captures fine-grained and continuous temporal relationships across adjacent frames. While stacking more layers increases the temporal receptive field and captures longer-range dependencies, the additional information may not significantly benefit the evaluated tasks.

However, the 3-layer STE consistently achieves strong performance, securing the best or second-best average performance. It also delivers the best performance in several individual benchmarks, such as VideoMME and MLVU. 
These findings suggest that while increasing the temporal receptive field alone may not guarantee substantial performance gains, carefully selecting the number of layers can yield consistently strong results for designing temporal modules.

\textbf{Temporal learning space.} We investigate how the location of temporal learning, either in the visual or semantic space, affects model performance. Temporal encoders can be inserted after the vision encoder and before the vision-language projector (learning in the visual space) or after the projector and before the LLM (learning in the semantic space). Tab.~\ref{tab:placement_ablation} summarizes results comparing these two choices during both the pretraining and SFT stages. Across both backbones and training stages, learning temporal information in the visual space consistently outperforms the semantic space. This suggests that temporal correlations in visual features are continuous and well-suited for learning through sliding windows, whereas temporal information in the semantic space may be discontinuous, owing to tokenized patches. While SFT narrows the performance gap, likely due to the LLM's improved capacity for processing semantic temporal relationships, the gap persists, particularly for LLaVA-Video (0.7\% lower average performance).

Additionally, the embedding size in the semantic space is expanded, requiring a larger STE with substantially more parameters (approximately 9.7x larger, $\sim$2.65M vs. $\sim$25.69M). This further underscores the inefficiency of learning temporal information in the semantic space. These findings highlight the importance of applying convolutional temporal modeling in the visual space, where temporal relationships are more continuous and computationally efficient.

\textbf{Stacking strategy for STE.}  
We further analyze layer stacking strategies for STE in Appendix~\ref{appendix:stacking}, exploring different combinations of learning spaces and compression ratios.

\subsection{Broader implications}

In this section, we explore the broader implications of explicit temporal modeling, particularly when using STE as a plug-in module and in image modalities.

\textbf{Compression Effectiveness as a Plug-in Module.}  
We examine STE’s utility as a plug-in module for video understanding tasks, focusing on its ability to compress temporal frames while maintaining performance. Tab.~\ref{plug_in_exp} presents the results of using STE pre-trained on video data and evaluated under varying frame reductions. We stack 1 to 4 STE layers with a (2:1) I/O frame ratio, progressively reducing frames by 50\%, 75\%, 87.5\%, and 93.75\%.

Despite being lightweight (fewer than 5.31M trainable parameters compared to the 7B backbone), STE demonstrates robust performance. With the full set of frames, STE improves LLaVA-Video's accuracy by 0.4\% and maintains LLaVA-OV's performance. Even under significant frame reductions, STE effectively mitigates performance degradation. For instance, with a 75\% frame compression, LLaVA-OV experiences only a 1.9\% drop in average accuracy and LLaVA-Video drops by just 2.2\%. Remarkably, even with a 93.75\% frame compression, the models achieve 50.3\% (LLaVA-OV-STE) and 60.6\% (LLaVA-Video-STE) average accuracy, demonstrating STE’s robustness under extreme compression.
These findings highlight STE’s ability to compensate for information loss caused by frame reductions by capturing longer-range temporal dependencies, thereby slowing performance decline. This underscores STE’s broader applicability as a plug-in module for fast adaptation without SFT to video datasets and for computationally constrained scenarios requiring frame reduction.

\textbf{Performance for image modalities.}
As shown in Tab.~\ref{tab:image_ability}, we calculate the mean relative score to the original LLaVA-OV using the formula \(\frac{1}{N} \sum_{i=1}^{N} \frac{\text{score}_{\text{STE},i}}{\text{score}_{\text{OV},i}}\) to evaluate the model's image ability after SFT. Since our pretraining only updates our module, which does not apply to images, the model's image ability changes only after SFT. Therefore, we investigate the changes in image ability after fine-tuning the entire model with video data. We evaluate our model on 9 single-image benchmarks and 1 multi-image benchmark (i.e., MuirBench). From the table, we observe that while the model's performance improves on the multi-image benchmark, it decreases on most single-image benchmarks. In most cases, our model maintains around 95\% of the original image ability, indicating that training the model exclusively on video data results in a slight decrease in image ability and emphasizes the need for image data during video training.

\textbf{Temporal receptive field sensitivity in image ability:} Stacking additional layers with (2:2) and (2:1) results in a continuous decline in image ability, with the decrease becoming more pronounced. This suggests that visual embeddings encoded with longer temporal information exacerbate the gap between video and image representations.

\section{Conclusion}

In this work, we systematically investigate the role of explicit temporal modeling in MLLMs for video understanding. We propose STE, designed to explicitly model temporal information with adjustable receptive fields and frame compression. Experiments show that STE enhances overall performance and temporal understanding across benchmarks, highlighting the importance of explicit temporal modeling in video MLLMs. We analyze its key design factors, such as its placement in the MLLM pipeline and robustness to temporal receptive field changes. We demonstrate its practical advantages, including efficient frame compression and adaptability as a plug-in module, and acknowledge its limitations in image modalities. These findings underscore the value of explicit temporal modeling and offer insights for advancing video MLLM design.

\section*{Acknowledgements}
We would like to express our gratitude to Jiaxian Guo from Google Research, Australia, for his valuable contributions to this project. We appreciate his efforts and insightful discussions during our work with TikTok.

\bibliography{example_paper}
\bibliographystyle{icml2025}

\newpage
\appendix
\onecolumn

\begin{figure*}[!t]
    \centering
    \begin{subfigure}{0.33\textwidth} 
        \includegraphics[width=\textwidth]{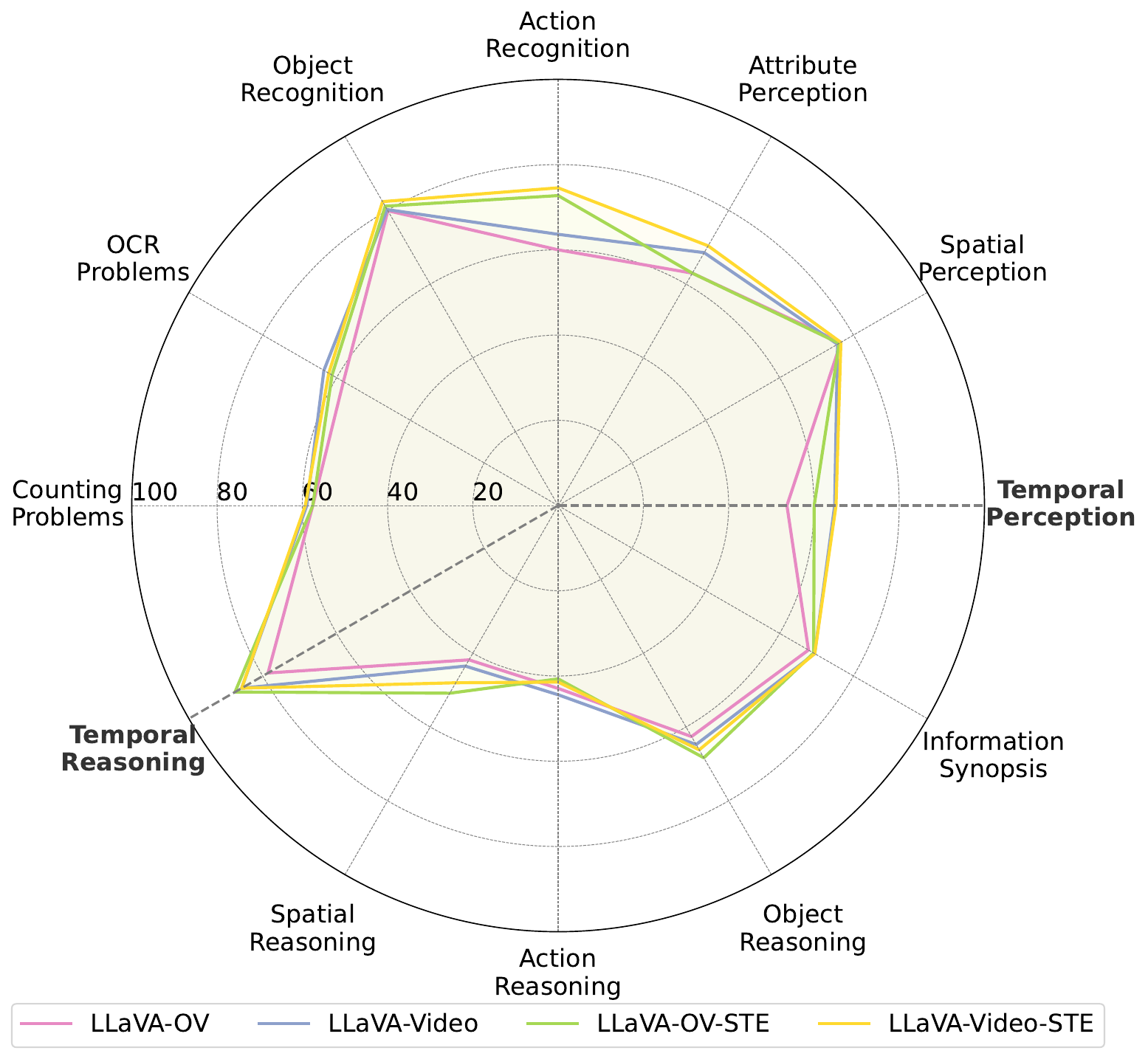}
        \caption{VideoMME with subtitle.}
    \end{subfigure}
    \hfill
    \begin{subfigure}{0.33\textwidth}
        \includegraphics[width=\textwidth]{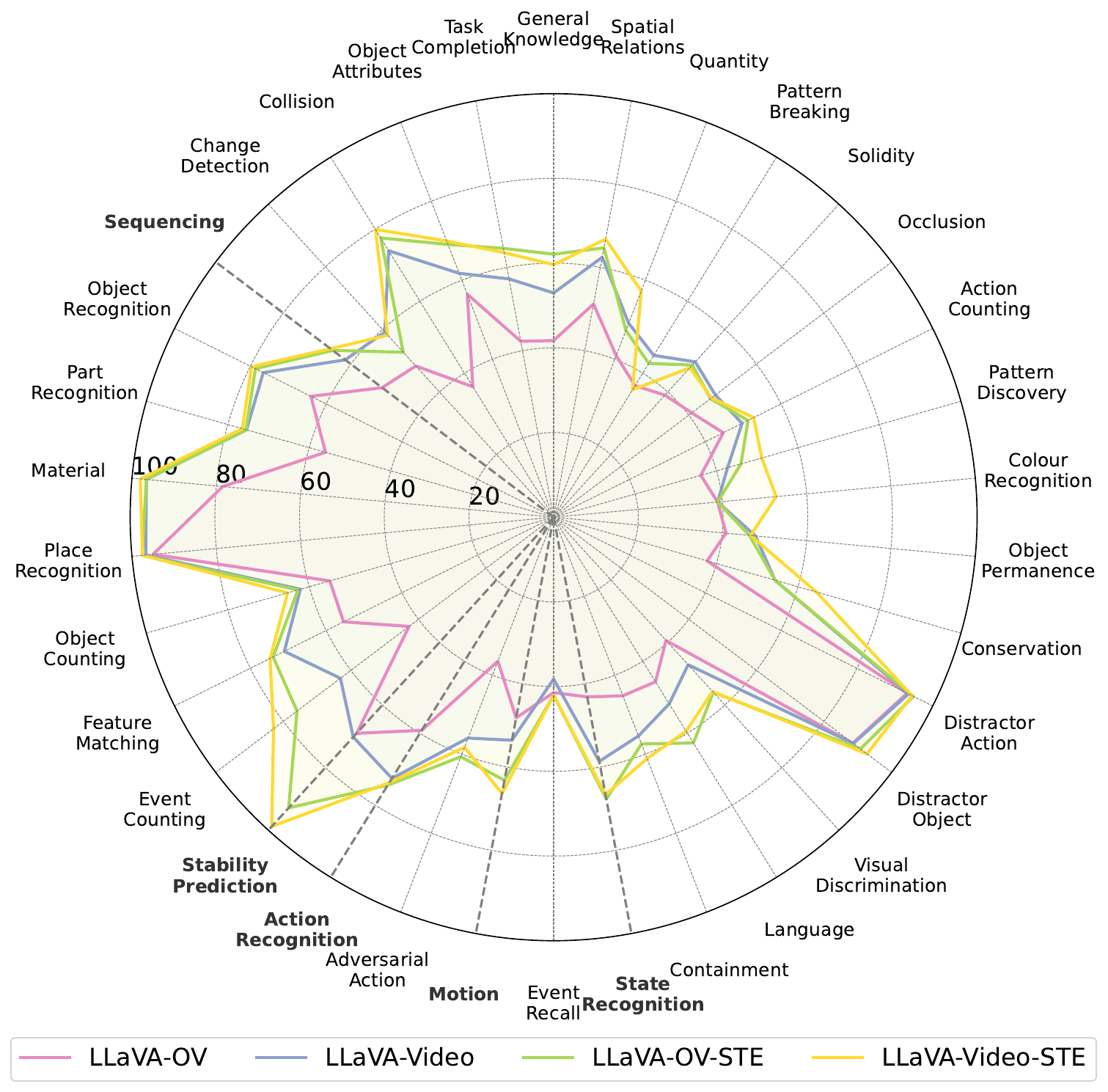}
        \caption{PerceptionTest.}
    \end{subfigure}
     \hfill
    \begin{subfigure}{0.33\textwidth}
        \includegraphics[width=\textwidth]{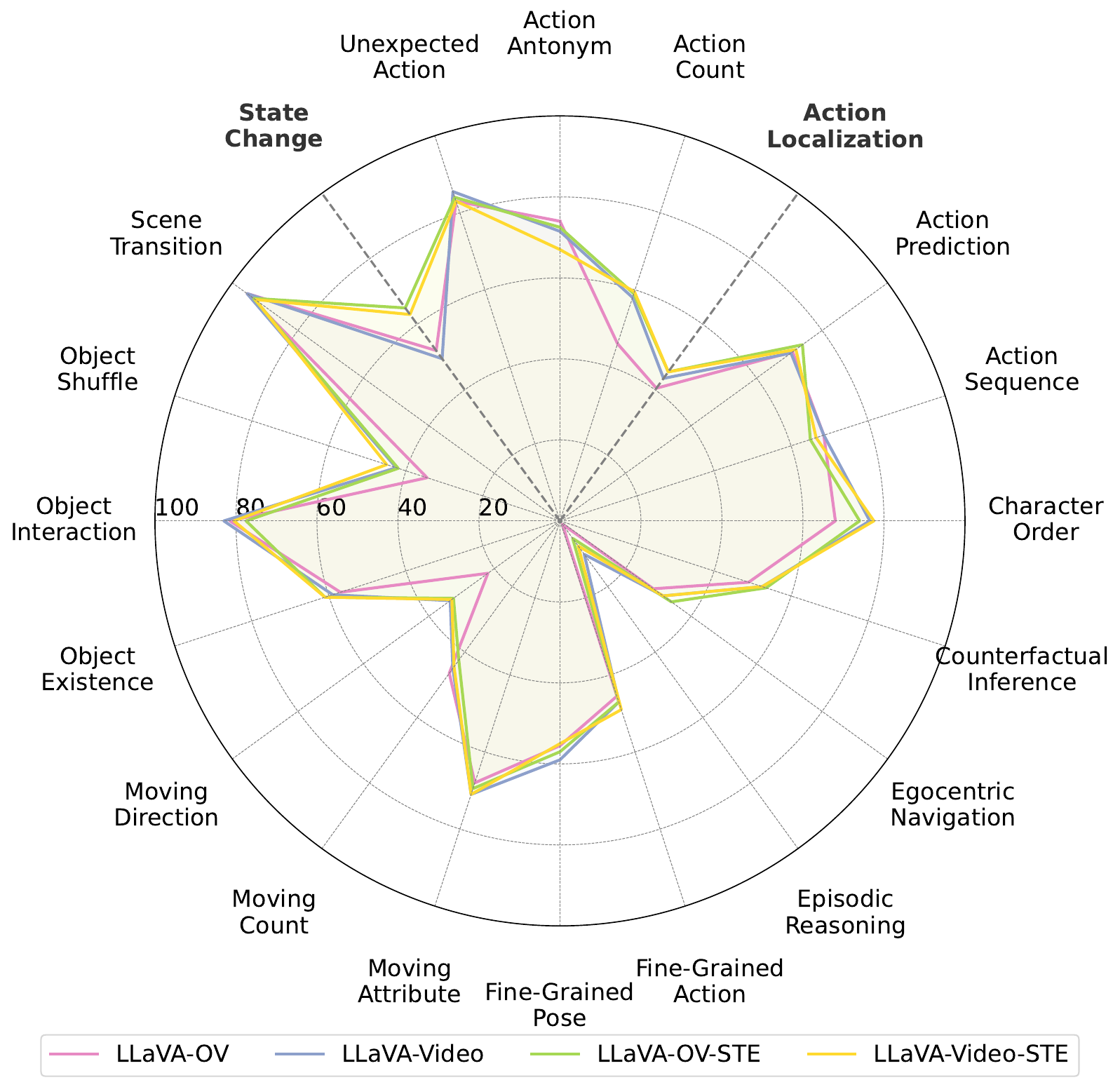}
        \caption{MVBench.}
    \end{subfigure}
    \vspace{1em}
    \begin{subfigure}{0.33\textwidth}
        \includegraphics[width=\textwidth]{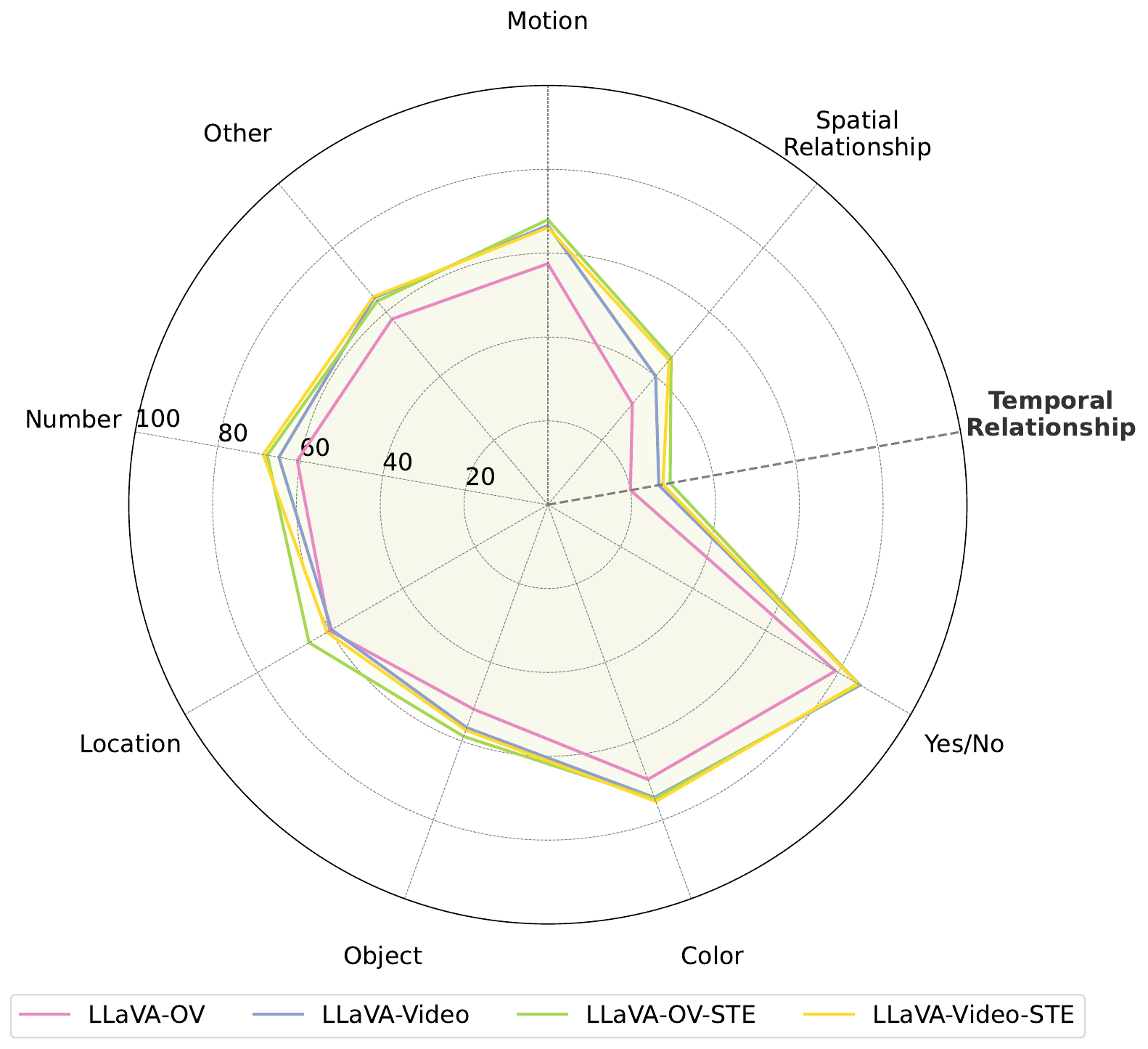}
        \caption{ActNet-QA.}
    \end{subfigure}
    \hfill
    \begin{subfigure}{0.33\textwidth}
        \includegraphics[width=\textwidth]{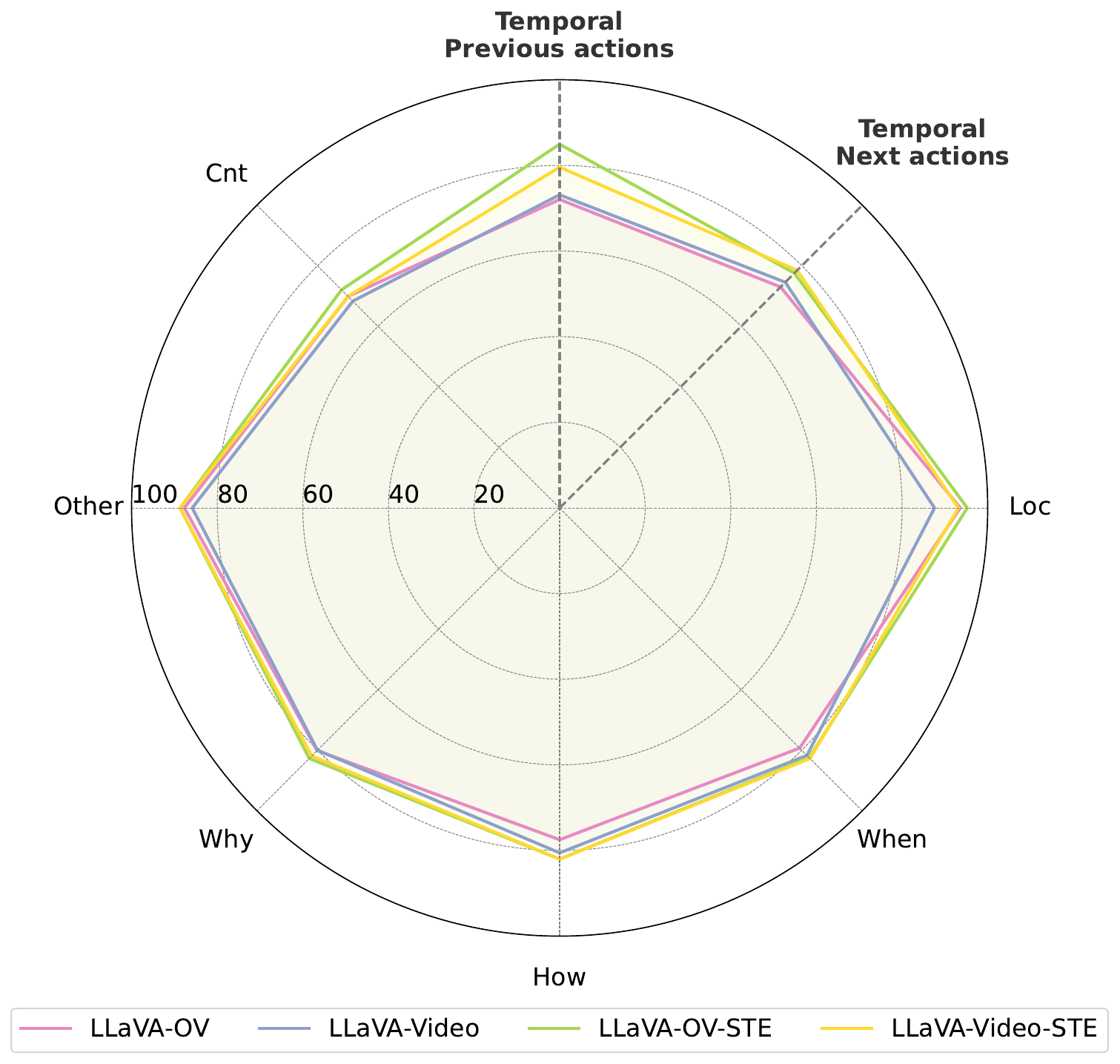}
        \caption{NExT-QA.}
    \end{subfigure}
     \hfill
     \begin{subfigure}{0.33\textwidth}
        \includegraphics[width=\textwidth]{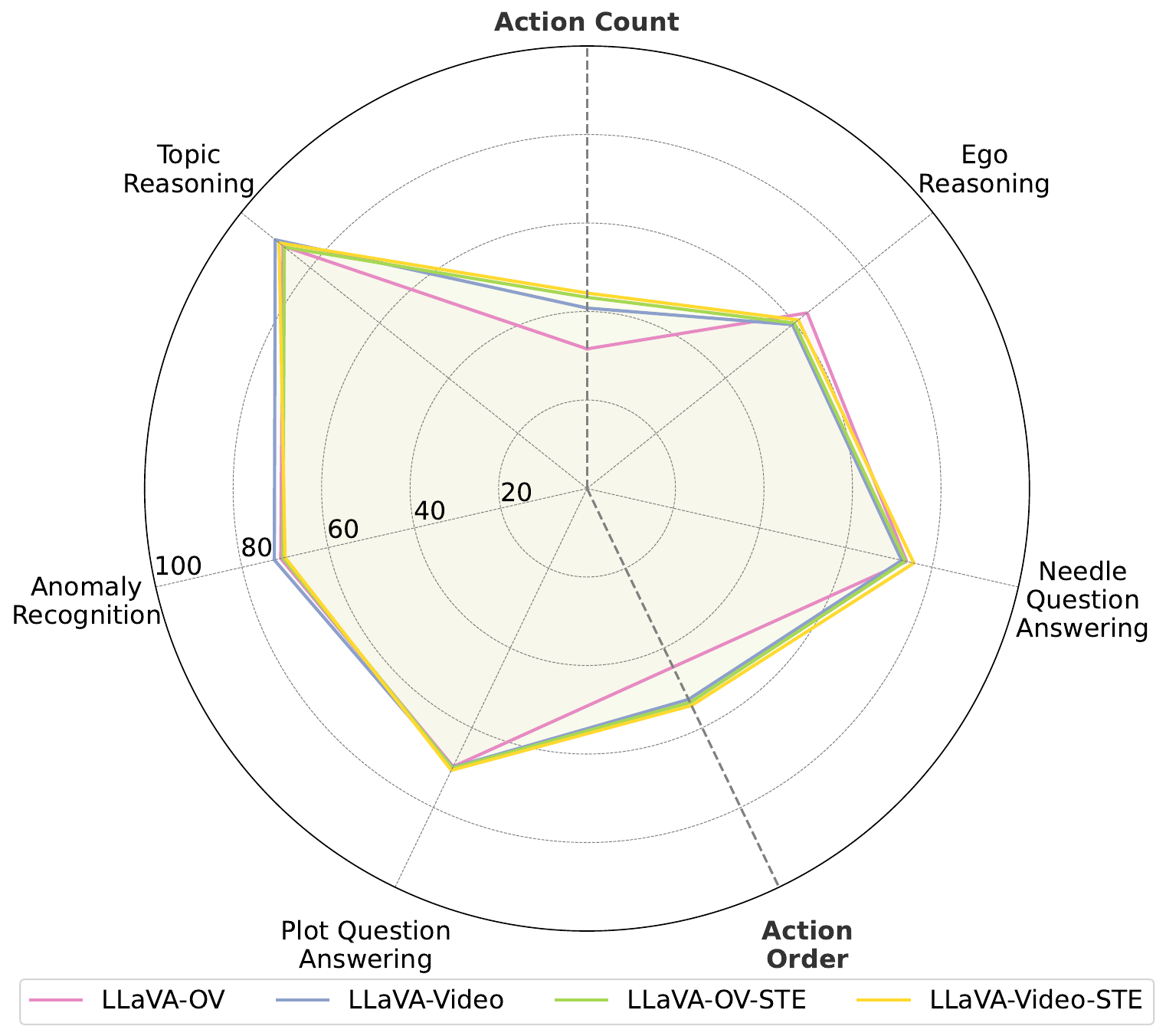}
        \caption{MLVU.}
    \end{subfigure}
    \caption{Task-level performance on benchmarks. LLaVA-OV-STE and LLaVA-Video-STE refer to LLaVA-OV-STE-3-(2:2) and LLaVA-Video-STE-3-(2:2), respectively.)}
    \label{fig:task_radar}
\end{figure*}

\begin{figure*}[!t]
    \centering
    \begin{subfigure}{0.9\textwidth}
        \centering
        \includegraphics[width=\textwidth]{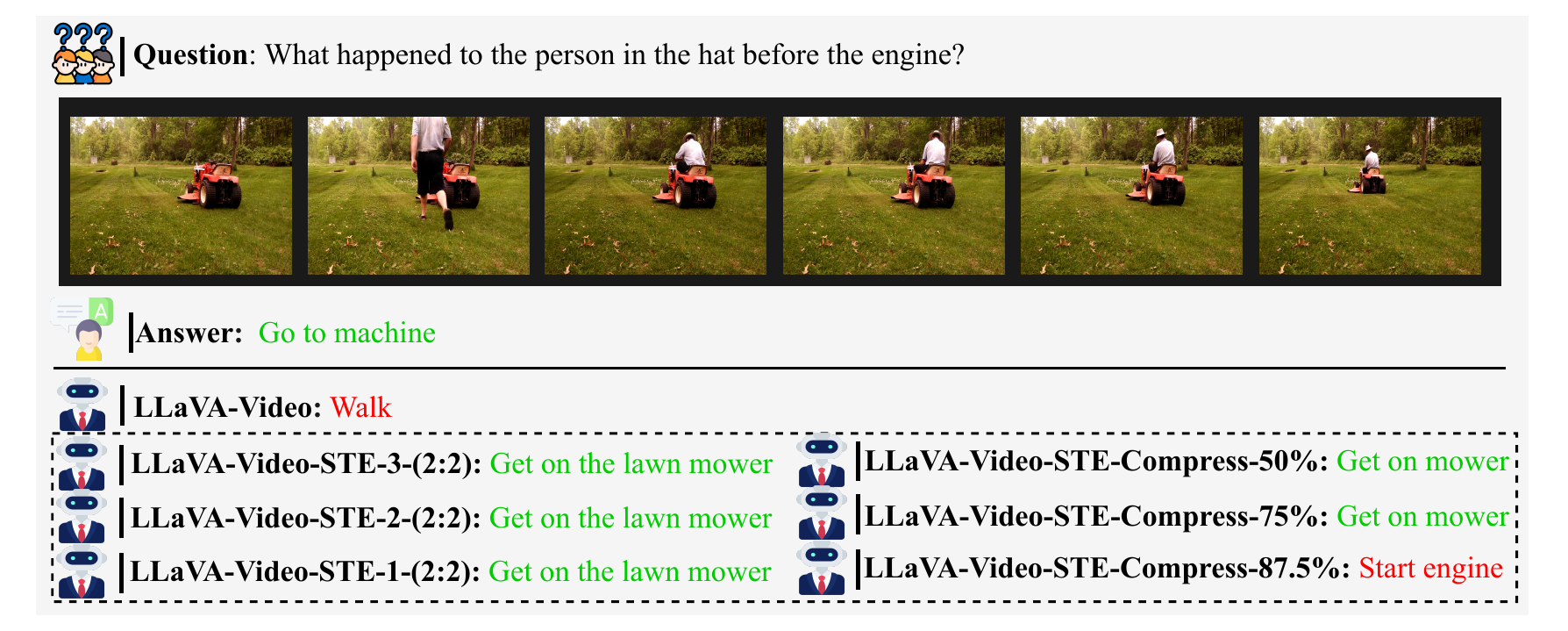}
        \caption{Case of temporal relationship task.}
    \end{subfigure}%
    \\
    \begin{subfigure}{0.9\textwidth}
        \centering
        \includegraphics[width=\textwidth]{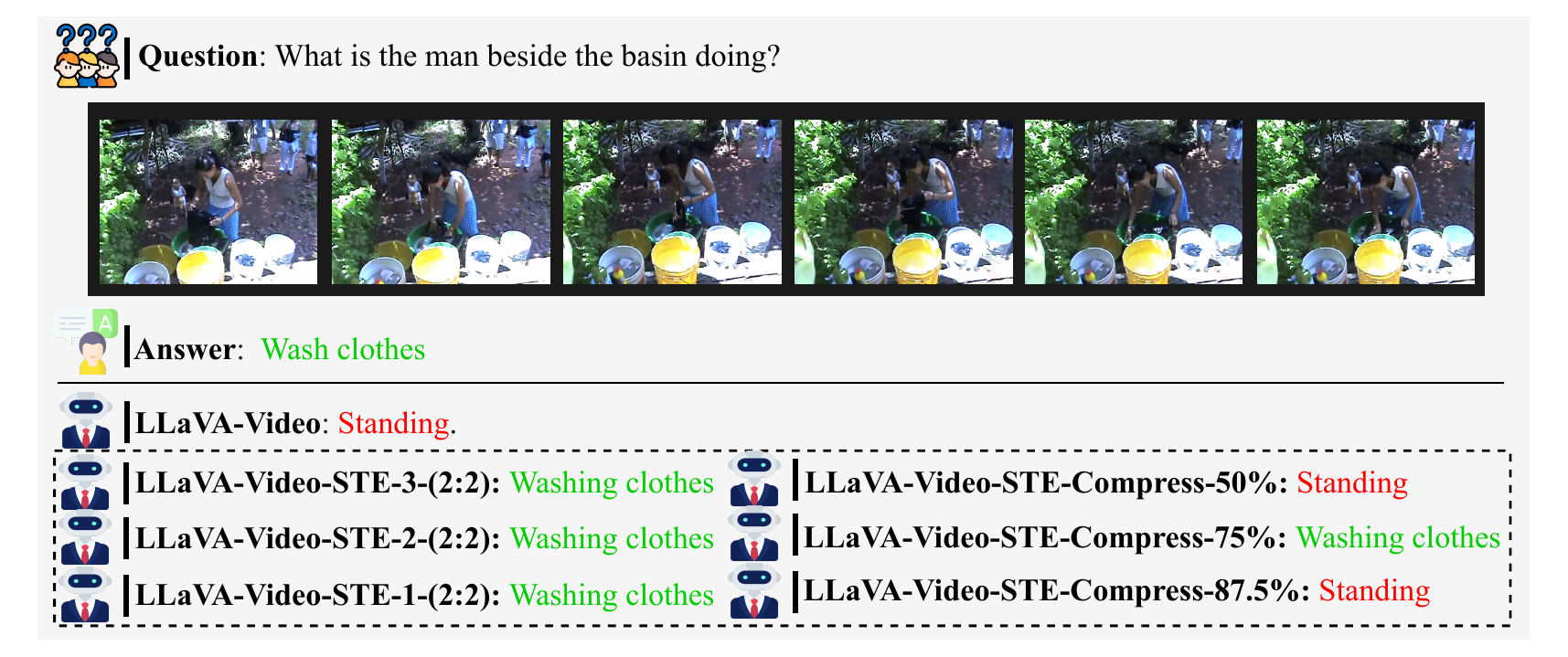}
        \caption{Case of motion task.}
    \end{subfigure}
    \caption{ActNet-QA case study. LLaVA-Video-STE-1/2/3-(2:2) represents LLaVA-Video-STE using 1/2/3 layers of (2:2).}
    \label{fig:case_study}
\end{figure*}

\section{Experiment Setting details}\label{exp_set_details}
For all STE layers, we use unified hyperparameters for the sliding window size and stride, setting \((T_{w}=2, T_{s}=1)\). This configuration is chosen to avoid significant padding when using large sliding windows (e.g., a 16-frame window with a stride of 1 requires padding up to 15 frames), which is inefficient and may impair tasks like occurrence counting. By adopting small sliding windows and strides, we minimize padding requirements while capturing finer temporal information. Furthermore, we stack multiple convolutional layers to gradually expand the temporal receptive field and explore the influence of different temporal learning scales.

All experiments are conducted on a cluster of 48 NVIDIA H100 GPUs. For both training and evaluation, 32 frames are sampled from each video. The model is trained for one epoch during both the pretraining and SFT phases. The batch size per GPU is set to 2, except for SFT with STE layers configured as \((T_{u}\):\(T_{o})\)=(2:2), where a batch size of 1 is used to prevent out-of-memory errors.

\section{Training Dataset Details}\label{training_dataset_details}
We utilize the following datasets for training:

\noindent\textbf{ActNet-QA}~\cite{yu2019activitynet}: A dataset designed for activity-based video question answering, containing 23,530 open-ended QA items.

\noindent\textbf{NExT-QA}~\cite{xiao2021next}: A dataset supporting temporal and causal relation reasoning in videos, with 17,090 open-ended QA items and 17,024 multiple-choice QA items.

\noindent\textbf{PerceptionTest}~\cite{patraucean2023perception}: A dataset targeting fundamental perceptual understanding of videos, comprising 1,803 open-ended QA items.

\noindent\textbf{LLaVA-Hound}~\cite{zhang2024direct}: A diverse video comprehension dataset with 240,000 open-ended QA items and 15,000 caption entries.

\noindent\textbf{LLaVA-Video-178K}~\cite{zhang2024video}: A comprehensive benchmark featuring videos of 0–3 minutes in duration, including 178,510 caption entries, 960,792 open-ended QA items, and 196,198 multiple-choice QA items.

\section{Task-level Ability Across Benchmarks}\label{task-level-ability}

We show the task-level model performance in Fig.~\ref{fig:task_radar}.

\textbf{VideoMME}: STE consistently enhances models' abilities in temporal reasoning, temporal perception, spatial reasoning, and action recognition, while other abilities remain at similar performance levels.

\textbf{PerceptionTest}: STE significantly improves models' performance in stability prediction, recognition of actions, object attributes, task completion, state, and motion, with other abilities showing comparable performance.

\textbf{MVBench}: STE consistently boosts models' capabilities in state change and action localization, while other abilities are unaffected.

\textbf{ActNet-QA}: STE consistently enhances models' capabilities in temporal relationship and spatial relationship, while other abilities are unaffected.

\textbf{NExT-QA}: STE improves models' understanding of temporal sequences, particularly in predicting previous and next actions, with no significant changes in other abilities.

\textbf{MLVU}: STE enhances models' proficiency in action count and order, with a slight improvement in ego reasoning. Other abilities remain consistent.

\textbf{Summary}: Across diverse video benchmarks, STE consistently strengthens models' temporal and spatial reasoning capabilities, particularly excelling in action recognition, state change, temporal sequence prediction, and stability analysis. While other abilities remain largely unaffected, these results underscore the value of explicit temporal modeling in enhancing models' understanding of complex temporal and spatial relationships in video data.

\section{Qualitative results}\label{Qualitative}

As shown in Fig.~\ref{fig:case_study}, we compare LLaVA-Video with and without our module on a temporal relationship task and a motion recognition task from ActNet-QA. We find that LLaVA-Video often relies on single-frame understanding, neglecting temporal context, e.g., identifying ``walk" or ``standing" without considering temporal information. In contrast, our module incorporates a broader temporal context, enabling more accurate predictions. For example, in Fig.~\ref{fig:case_study}(a), it integrates the surrounding frames to deduce that the man is to ``get on the lawn mower." However, reducing tokens can lead to missed information, e.g., predicting ``mower" instead of ``lawn mower." Interestingly, when 87.5\% of tokens are compressed, the model focuses on high-level semantics, predicting that the reason for the man going to the machine is to start the engine. Fig.~\ref{fig:case_study}(b) highlights another key result: longer temporal encoding may compensate for missed information caused by token reduction, enabling correct recognition of ``washing" with a 75\% token reduction. This shows that an extended temporal receptive field can mitigate information loss under high-ratio token compression.

\begin{table}[t]
\centering
\caption{Learning space ablation as a plug-in module.}
\label{tab:vision_token_ablation}
\resizebox{0.65\linewidth}{!}{
\begin{tabular}{c|c|cc}
\toprule
\multirow{2}{*}{Before Projector} & \multirow{2}{*}{After Projector} & VideoMME & VideoMME \\
 &  & (w/sub) & (w/o sub) \\
\midrule
(4:3) & $\times$ & 56.3 & 58.9 \\
$\times$ & (4:3) & 54.3 & 58.3 \\
 \midrule
(2:1) & $\times$ & 57.6 & 59.8 \\
$\times$ & (2:1) & 52.8 & 57.5 \\
 \midrule
(2:1)-(2:1) & $\times$ & 55.7 & 58.5 \\
$\times$ & (2:1)-(2:1) & 54.6 & 58.2 \\
(2:1) & (2:1) & 50.7 & 56.3 \\
\bottomrule
\end{tabular}
}
\end{table}

\section{Stacking strategies for temporal encoders.} \label{appendix:stacking} 
We further analyze stacking strategies for STE under varying compression ratios and temporal learning spaces. Tab.~\ref{tab:vision_token_ablation} presents the results. When using STE as a 1-layer plug-in, applying convolution in the semantic space results in a greater performance drop as the compression ratio increases. For example, on VideoMME without subtitles, the performance drops by 2\% and 4.8\% for semantic-space convolution compared to visual-space convolution under different compression ratios. When comparing three stacking strategies (visual space only, semantic space only, and both spaces), applying convolution exclusively in the visual space consistently achieves the best performance. Both visual-only and semantic-only configurations significantly outperform the combined approach (convolution in both spaces). The combined approach introduces higher computational overhead and requires more training, making it less efficient. These results emphasize the importance of domain-specific modeling, with the visual space offering the best performance and efficiency.

\begin{table*}[t]
\centering
\caption{Dataset ablation of LLaVA-OV. }
\label{tab:dataset_ablation}
\resizebox{\textwidth}{!}{%
\begin{tabular}{c|c|c|ccccccc|c}
\toprule
 \multirow{2}{*}{STE}           & \multirow{2}{*}{Training Phase}       & \multirow{2}{*}{Dataset}                   & \multirow{2}{*}{PerceptionTest} & \multirow{2}{*}{ActNet-QA} & \multirow{2}{*}{NExT-QA} & \multirow{2}{*}{MLVU} &\multirow{2}{*}{MVBench}& VideoMME & VideoMME & \multirow{2}{*}{AVG}   \\
                         & & & & & &  && (w/o sub) & (w/sub) &   \\
                         \midrule
 &                            & All five datasets         & 58.6           & 58.7      & 79.6   & 65.1 &55.0& 58.9              & 60.0            & 62.3\\
 &                            & Remove LLaVA-Hound        & 59.3           & 58.3      & 79.6   & 64.5 &54.9& 58.1              & 60.2            & 62.1\\
 & \multirow{-3}{*}{Pretrain} & Only use LLaVA-Video-178K & 58.3           & 58.2      & 79.6   & 65.7 &54.6& 58.4              & 60.9            & 62.2\\
 \cline{2-11}
 &                            & All five datasets         & 70.1           & 65.7      & 82.4   & 66.9 &57.8& 60.0              & 63.1            & 66.6\\
 &                            & Remove LLaVA-Hound        & 71.0           & 65.2      & 83.2   & 66.9 &58.2& 61.5              & 63.6            & 67.1\\
\multirow{-6}{*}{\(+\) (2:2)}      & \multirow{-3}{*}{SFT}      & Only use LLaVA-Video-178K & 56.2           & 54.4      & 66.4   & 66.3 &52.5& 58.1              & 60.2            & 59.2\\
\midrule
 &                            & All five datasets         & 57.9           & 58.9      & 78.9   & 63.8 &53.9& 57.6              & 59.8            & 61.5\\
 &                            & Remove LLaVA-Hound        & 57.7           & 58.1      & 79.3   & 64.2 &54.0& 56.9              & 59.4            & 61.4\\
 & \multirow{-3}{*}{Pretrain} & Only use LLaVA-Video-178K & 57.2           & 57.5      & 79.1   & 63.6 &53.2& 57.0              & 59.3            & 61.0\\
 \cline{2-11}
 &                            & All five datasets         & 69.4           & 60.1      & 79.1   & 66.9 &55.3& 60.4              & 63.2            & 64.9\\
 &                            & Remove LLaVA-Hound        & 68.1           & 59.5      & 78.9   & 67.4 &55.2& 60.1              & 63.3            & 64.6\\
\multirow{-6}{*}{\(+\) (2:1)}     & \multirow{-3}{*}{SFT}      & Only use LLaVA-Video-178K & 54.8           & 57.5      & 69.7   & 64.8 &50.3& 57.5              & 59.8            & 59.2\\
\midrule
 &                            & All five datasets         & 56.9           & 57.2      & 78.1   & 63.0 &53.6& 55.7              & 58.5            & 60.4\\
 &                            & Remove LLaVA-Hound        & 57.4           & 57.5      & 76.9   & 62.3 &52.7& 55.3              & 58.4            & 60.1\\
 & \multirow{-3}{*}{Pretrain} & Only use LLaVA-Video-178K & 56.5           & 57.7      & 77.2   & 63.1 &52.1& 54.9              & 57.5            & 59.9\\
 \cline{2-11}
 &                            & All five datasets         & 67.9           & 59.7      & 79.0   & 66.2 &54.8& 59.4              & 62.4            & 64.2\\
 &                            & Remove LLaVA-Hound        & 69.0           & 58.8      & 79.0   & 65.7 &55.6& 59.6              & 62.5            & 64.3\\
\multirow{-6}{*}{\(+\) (2:1)-(2:1)} & \multirow{-3}{*}{SFT}      & Only use LLaVA-Video-178K & 54.7           & 55.5      & 68.2   & 66.3 &51.3& 56.8              & 60.3            & 59.0\\
\bottomrule
\end{tabular}%
}
\end{table*}

\section{Dataset Ablation}

As shown in Tab.~\ref{tab:dataset_ablation}, we investigate the impact of dataset selection by considering 1-layer STE of (2:2) and 1/2-layer STE of (2:1) to assess the influence of datasets on maintaining or compressing frame counts. We conduct experiments based on LLaVA-OV, following the dataset split settings from \cite{zhang2024video}. Our analysis reveals several key insights. First, during the pretraining stage, using all five datasets consistently yields the best performance. The use of only LLaVA-Video-178K or LLaVA-Video-178K combined with three additional in-domain datasets does not significantly impact the final results. Second, during the SFT stage, we continue fine-tuning the model with the same datasets used in pretraining. The results indicate that whether LLaVA-hound is included or not, the average score remains similar. However, using only LLaVA-Video-178K leads to a significant decrease in performance across all benchmarks. Based on these findings, we recommend using all five datasets for stable pretraining and SFT performance in video training.


\end{document}